\documentclass[lettersize,journal]{IEEEtran}
\usepackage{cite}
\usepackage{amsmath,amssymb,amsfonts}
\usepackage{algorithmic}
\usepackage{graphicx}
\usepackage{textcomp}

\usepackage{dsfont}
\usepackage{multicol}
\usepackage{subcaption}
\usepackage{url}
\usepackage{booktabs}
\usepackage{multirow}
\usepackage{xcolor}
\usepackage{color}

\newcommand{\blue}[1]{\textcolor{blue}{#1}}
\renewcommand{\blue}[1]{#1}
\newcommand{\bluee}[1]{\textcolor{blue}{#1}}
\renewcommand{\bluee}[1]{#1}
\newcommand{\stdev}[1]{{\tiny $\pm$#1}}

\begin{document}

\title{Unleashing the Power of Intermediate Domains for Mixed Domain Semi-Supervised Medical Image Segmentation}
\author{Qinghe Ma, Jian Zhang, Lei Qi, Qian Yu, Yinghuan Shi, Yang Gao
\thanks{This work was supported by National Science and Technology Major Project (2023ZD0120700), NSFC Project (62222604, 62206052, 624B2063), China Postdoctoral Science Foundation (2024M750424), Fundamental Research Funds for the Central Universities (020214380120, 020214380128), State Key Laboratory Fund (ZZKT2024A14, ZZKT2025B05), Postdoctoral Fellowship Program of CPSF (GZC20240252), Jiangsu Funding Program for Excellent Postdoctoral Talent (2024ZB242), Jiangsu Science and Technology Major Project (BG2024031), and Shandong Natural Science Foundation (ZR2023MF037)}
\thanks{Qinghe Ma, Jian Zhang, Yinghuan Shi, and Yang Gao are with the State Key Laboratory for Novel Software Technology, Nanjing University, China. They are also with National Institute of Healthcare Data Science, Nanjing University, China. Yinghuan Shi is also with Nanjing Drum Tower Hospital, Nanjing, Jiangsu, China. (E-mail: mqh@smail.nju.edu.cn, zhangjian7369@smail.nju.edu.cn, syh@nju.edu.cn, gaoy@nju.edu.cn)}
\thanks{Lei Qi is with the School of Computer Science and Engineering, and the Key Lab of Computer Network and Information Integration (Ministry of Education), Southeast University, China. (E-mail: qilei@seu.edu.cn)}
\thanks{Qian Yu is with the School of Data and Computer Science, ShandongWomen’s University, China. (E-mail: yuqian@sdwu.edu.cn)}
\thanks{The corresponding author of this work is Yinghuan Shi.}
}

\maketitle

\begin{abstract}
Both limited annotation and domain shift are prevalent challenges in medical image segmentation. Traditional semi-supervised segmentation and unsupervised domain adaptation methods address one of these issues separately. However, the coexistence of limited annotation and domain shift is quite common, which motivates us to introduce a novel and challenging scenario: \textbf{Mi}xed \textbf{D}omain \textbf{S}emi-supervised medical image \textbf{S}egmentation (MiDSS), where limited labeled data from a single domain and a large amount of unlabeled data from multiple domains.
To tackle this issue, we propose the UST-RUN framework, which fully leverages intermediate domain information to facilitate knowledge transfer.
We employ Unified Copy-paste (UCP) to construct intermediate domains, and propose a Symmetric GuiDance training strategy (SymGD) to supervise unlabeled data by merging pseudo-labels from intermediate samples. Subsequently, we introduce a Training Process aware Random Amplitude MixUp (TP-RAM) to progressively incorporate style-transition components into intermediate samples. 
To generate more diverse intermediate samples, we further select reliable samples with high-quality pseudo-labels, which are then mixed with other unlabeled data.
Additionally, we generate sophisticated intermediate samples with high-quality pseudo-labels for unreliable samples, ensuring effective knowledge transfer for them. 
Extensive experiments on four public datasets demonstrate the superiority of UST-RUN. Notably, UST-RUN achieves a 12.94\% improvement in Dice score on the Prostate dataset.
Our code is available at \textcolor{magenta}{\url{https://github.com/MQinghe/UST-RUN}}.
\end{abstract}

\section{Introduction}
\label{intro}

\begin{figure}[!t]
\centering
\includegraphics[width=1.0\columnwidth]{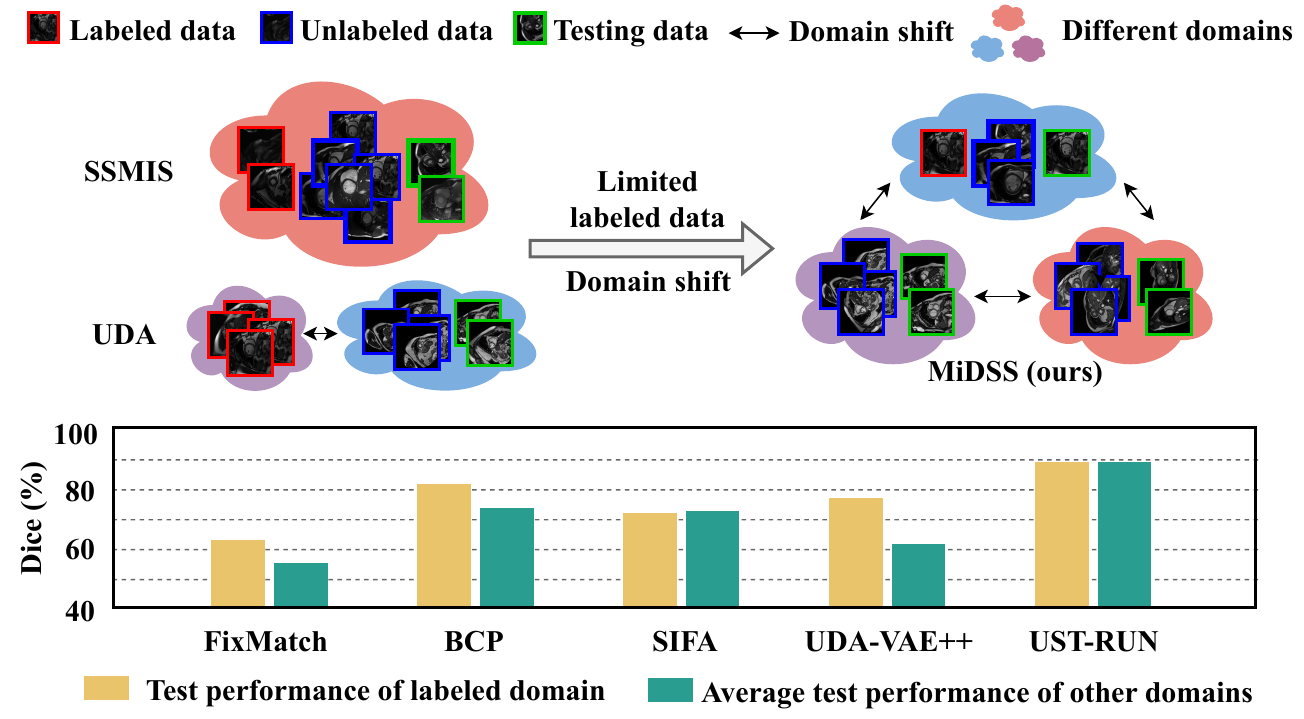}
\caption{The upper illustrates SSIMS, UDA, and MiDSS. The lower shows the comparison between different methods on the labeled domain (BIDMC)~\cite{liu2020shape} and other domains.}
\label{setting}
\end{figure}

Semi-supervised medical image segmentation (SSIMS) has gained significant attention in recent years for its advantage to enhance performance while alleviating the labeling burden~\cite{yu2019uncertainty, fan2020inf, li2020transformation,zeng2023ss}. 
By utilizing only a few labeled data and leveraging advancements in unsupervised learning techniques, \textit{e.g.}, contrastive learning~\cite{he2020momentum}, SSIMS has become a promising approach to effectively exploit unlabeled data. Traditional SSIMS methods generally assume that labeled and unlabeled data originate from the same distribution, ensuring consistency during training.

Despite their success, a critical yet often overlooked issue arises:
\emph{During clinical data collection, do we really check and guarantee, whether each unlabeled sample belongs to the same distribution as labeled samples?}  
For instance, while a certain amount of labeled data is carefully collected and annotated from a specific center, unlabeled data are often more easily accessible in large quantities. In such scenarios, physicians may lack sufficient time or resources to thoroughly assess whether these unlabeled data share the same distribution as the labeled ones~\cite{dash2019big, rudrapatna2020opportunities}.

As noted earlier, the dilemma of ``domain shift"~\cite{wang2019semi,guan2021domain} can severely degrade model performance, posing a significant challenge when developing a generalized SSIMS model. 
A more practical setting could be: given labeled data from a consistent distribution, the unlabeled data is allowed to come from the same distribution or 
multiple different distributions.  
It is worth noting that this challenge cannot be directly addressed by existing unsupervised domain adaptation (UDA) methods~\cite{chen2020unsupervised,russo2018source,dou2018unsupervised}. Although UDA holds promise for addressing domain shift, they still rely on abundant labeled data. 

Despite the significance of domain shift in SSIMS, according to our best knowledge, our work is first to attempt to investigate this issue, which we term \textbf{Mi}xed \textbf{D}omains \textbf{S}emi-supervised medical image \textbf{S}egmentation (abbreviated as \textbf{MiDSS}). Imaging in clinical applications, as illustrated in Fig.~\ref{setting}, given numerous unlabeled data from multiple centers, how can we train a model with only a few labeled samples from a typical center to perform well across all these centers? 
Furthermore, existing SSMIS and UDA methods are not well-suited for this scenario, as they primarily focus on either addressing limited annotations or mitigating domain shifts independently. This limitation underscores the necessity of the MiDSS. 
Compared to semi-supervised domain adaptation (SSDA)~\cite{li2021ecacl,kim2020attract,yan2022multi}, which also deals with limited annotation and domain shift, MiDSS poses a more complex scenario. In SSDA, a small number of labeled target data are available alongside labeled source data, with unlabeled data solely from a single target domain. In contrast, MiDSS relies on limited labeled data from a single source domain while leveraging unlabeled data from both the source domain and multiple diverse target domains, making adaptation more challenging.

The key to addressing the problem lies in \textit{how to generate reliable pseudo-labels for unlabeled data, especially in the presence of domain shift}. 
However, the domain gap across training data often leads to inaccurate and unreliable pseudo-labels, undermining the effectiveness of self-training.
Thus, the key insight is to first construct intermediate domains to reduce the domain gap and then enhance knowledge transfer across domains.
CutMix~\cite{yun2019cutmix}, also referred to as Copy-paste (CP), is a simple yet highly effective technique for generating intermediate samples by embedding patches from labeled samples into unlabeled samples along a single direction.
To enhance the data generation capability of CP, we employ Unified Copy-paste (UCP), which generates a great quantity of intermediate samples by unified applying both embedding directions between labeled and unlabeled samples, narrowing the domain gap and mitigating the issue of error accumulation. 

Intermediate domains do not inherently align with the distribution of unlabeled data, leading to potential discrepancies.
Focusing training solely on these domains risks overlooking the model performance on unlabeled data, which remains a key concern.
To mitigate this, information from intermediate domains should be effectively utilized to refine pseudo-labels for unlabeled data.
Moreover, medical images of the same organ or lesion often exhibit similar structures~\cite{wang2022separated}, with stylistic differences being the primary source of domain shift~\cite{liu2021feddg}. Local semantic mixing methods (\textit{e.g.}, CP) fail to handle these differences, resulting in a lack of stylistic transition. 
Besides, an aggressive stylistic transition hinders the progressive formation of intermediate domains.

To effectively leverage intermediate domains and establish a robust and stable knowledge transfer process, this work introduces a Mixed Domains Semi-Supervised Medical Image Segmentation approach. In terms of training strategy, we design Symmetric Guidance training strategy (SymGD), which enforces bidirectional guidance between unlabeled data and intermediate samples. This dual-perspective integration of pseudo-labels enhances the precision of guidance. Concerning intermediate samples, we introduce a Training Process-Aware Random Amplitude MixUp module (TP-RAM) to promote smooth stylistic knowledge transfer. The conference version of our method is named UST, derived from the initials of its three core modules: \textbf{U}CP, \textbf{S}ymGD, and \textbf{T}P-RAM.

The original UST method transfers knowledge solely from labeled to unlabeled data. However, with labeled data confined to a single domain and extremely limited in quantity (\textit{e.g.}, 5 examples), the diversity of generated intermediate samples is insufficient.
Additionally, when the pseudo-labels of the source images are of low quality,
inaccurate pseudo-labels for intermediate samples result in negative transfer. To address this, we enhance diversity by selecting unlabeled data with high-quality pseudo-labels (reliable samples) to generate more representative intermediate samples. To achieve efficient knowledge transfer for unlabeled data with low-quality pseudo-labels (unreliable samples), we craft more refined intermediate samples with reliable pseudo-labels. 
By leveraging \textbf{R}eliable and \textbf{UN}reliable samples, we propose the UST-RUN method.
Our main contributions can be summarized as follows:
\begin{itemize}
    \item We investigate a new yet underestimated semi-supervised medical image segmentation setting (namely MiDSS).
    \item A novel SymGD training strategy based on UCP, promotes the training on the unlabeled data with intermediate domains information.
    \item A TP-RAM module to make domain knowledge transfer comprehensive and stable.
\end{itemize}

This work is built upon our conference version~\cite{ma2024constructing}, and the contributions of this version are summarized as follows:

\begin{itemize}
    \item To enhance the diversity of intermediate samples, we select reliable samples to generate intermediate samples, facilitating knowledge transfer across domains. 
    \item To mitigate the adverse impact of low-quality pseudo-labels, we introduce a targeted training strategy that refines intermediate samples, ensuring more stable and accurate knowledge transfer for challenging cases. 
    \item Compared to UST, UST-RUN achieves superior performance across all four datasets, demonstrating superior segmentation accuracy, particularly in scenarios with extremely limited labeled data.
\end{itemize}

Extensive experiments validate the effectiveness of our method on four public datasets and show pronounced advantages with fewer labeled data.
On the Prostate dataset, our method achieves a remarkable improvement of 12.94\% Dice compared with other state-of-the-art methods.

\section{Related Work}
\label{related_work}

\begin{figure*}[!t]
\centering
\includegraphics[width=0.96\linewidth]{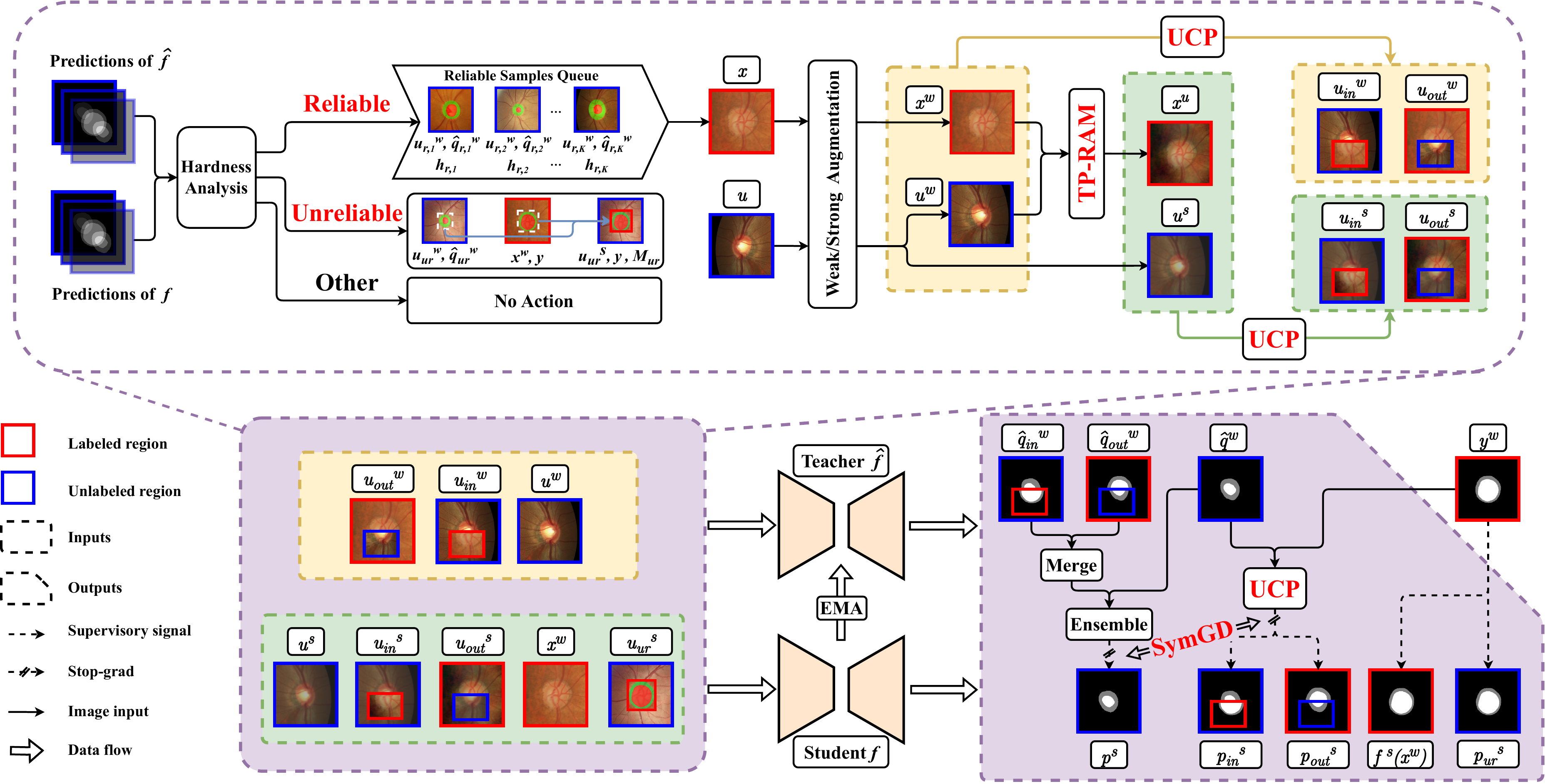}
\caption{The overall framework of our UST-RUN emphasizes domain knowledge transfer through data augmentation and training strategy. We generate intermediate samples through UCP between labeled data and unlabeled data. During training, we gradually introduce style transfer components to the intermediate samples, constructing intermediate domains at both semantic and stylistic levels. Moreover, we design a symmetric guidance training strategy, which operates between intermediate and unlabeled samples. In addition, we estimate the quality of pseudo-labels, selecting reliable samples to generate a more diverse set of intermediate samples. For unreliable samples, we focus on mitigating the adverse effects of low-quality pseudo-labels.}
\label{framework}
\end{figure*}

\subsection{Semi-supervised Medical Image Segmentation.}
Manually annotating medical images is challenging and costly~\cite{zhuang2013challenges}. Semi-supervised medical image segmentation methods have shown promise in addressing limited-label segmentation tasks. 
Entropy minimization and consistency regularization are widely adopted techniques in this context.
Yu \textit{et al.}~\cite{yu2019uncertainty} encourage consistent predictions under different perturbations. 
Li \textit{et al.}~\cite{li2020shape} introduce a shape-aware strategy using a multi-task deep network that jointly predicts semantic segmentation and signed distance maps. 
Luo \textit{et al.}~\cite{luo2021semi} propose a dual-task network that predicts pixel-wise segmentation maps and geometry-aware level set representations.
Wu \textit{et al.}~\cite{wu2022exploring} address challenges in SSIMS by simultaneously enforcing pixel-level smoothness and inter-class separation.
Miao \textit{et al.}~\cite{miao2023caussl} point out the importance of algorithmic independence between two networks in SSIMS. Chi \textit{et al.}~\cite{chi2024adaptive} propose an adaptive bidirectional displacement, which can employ multiple perturbations while guaranteeing the quality of consistency learning. Recently, foundation models~\cite{kirillov2023segment,zhang2023customized,chen2023sam,ma2024segment} have opened new avenues for semi-supervised medical image segmentation. Cheng \textit{et al.}~\cite{cheng2024unleashing} introduced a prompt-free adaptation of SAM, enabling efficient fine-tuning of medical images via a two-stage hierarchical decoding procedure. However, existing methods often struggle with domain shift, causing decreased performance due to the shared distribution assumption between labeled and unlabeled data.

\begin{table}[t]
  \centering
    \footnotesize
  \caption{\blue{Summary of notations.}}
\resizebox{\linewidth}{!}{
  \begin{tabular}{ll}
    \toprule
    \blue{Notation} & \blue{Definition}\\
    \midrule
    \blue{$\hat{f}$, $f$} & \blue{Teacher and student model}\\
    \blue{$x$, $x^w$, $x^u$} & \blue{Labeled data, weakly augmented $x$, style transition of $x$} \\
    \blue{$y$, $y^w$} & \blue{Ground truth of $x$, weakly augmented $y^w$} \\
    \blue{$u$, $u^w$, $u^s$} & \blue{Unlabeled data, weakly and strongly augmented $u$} \\
    \blue{$p^s$, $\hat{q}^w$} & \blue{Probability map on $u^s$ of $f$, pseudo label on $u^w$ of $\hat{f}$} \\
    \blue{$u_{in}^w$, $u_{out}^w$} & \blue{Intermediate sample from  $x^w$ and $u^w$} \\
    \blue{$u_{in}^s$, $u_{out}^s$} & \blue{Intermediate sample from  $x^u$ and $u^s$} \\
    \blue{$p_{in}^s$, $p_{out}^s$} & \blue{Probability map on $u_{in}^w$ and $u_{out}^w$ of $f$} \\
    \blue{$\hat{q}_{in}^w$, $\hat{q}_{out}^w$} & \blue{Pseudo label of $u_{in}^s$ and $u_{out}^s$ from $y^w$ and $\hat{q}^w$}\\
    \blue{$u_{r,i}^w$, $\hat{q}_{r,i}^w$, $h_{r,i}$} & \blue{The $i^{th}$ reliable sample and its pseudo label and hardness}\\
    \blue{$u_{ur}^w$, $\hat{q}_{ur}^w$} & \blue{Unreliable sample and its pseudo label}\\
    \blue{$M_{ur}$} & \blue{Merged foreground region of $\hat{q}_{ur}^w$ and $y^w$}\\
    \blue{$u_{ur}^s$, $p_{ur}^s$} & \blue{Intermediate sample from $u_{ur}^w$ and $x^w$, Probability map on $u_{ur}^s$ of $f$}\\
    \bottomrule
  \end{tabular}}
  \label{tab:notations}
\end{table}

\subsection{Unsupervised Domain Adaptation.} 
UDA methods train the model by leveraging abundant labeled source domain data and unlabeled target domain data to achieve appealing performance in the target domain. In medical image segmentation tasks, domain shift issues arise from variations in device parameters, disease severity, imaging principles, \textit{etc}. Thus, UDA plays a crucial role in addressing these issues. Adversarial learning based UDA
methods achieve alignment at multiple levels to narrow domain gap, including input alignment \cite{russo2018source,chartsias2017adversarial} and feature alignment \cite{ganin2016domain,dou2018unsupervised,zhao2022uda,tsai2018learning}. Self-training based UDA methods \cite{zheng2021rectifying, zhang2020collaborative} train the model by generating reliable pseudo-labels for unlabeled data. Methods like \cite{chen2022deliberated,yang2020fda} promote domain knowledge transfer by generating intermediate domains. These methods typically rely on a sufficient amount of source domain data and focus on a single target domain, making it challenging to be effective in MiDSS. As for test-time adaptation methods~\cite{wang2020tent, chen2024each}, which aim to adapt a model pre-trained on the source domain to the test data in a source-free and online manner, they require a large number of labeled data to ensure strong performance on the source domain. However, in the MiDSS scenario, the limited amount of labeled data fails to meet this requirement.

\subsection{\blue{Semi-supervised domain adaptation}}
\blue{Compared to UDA, where there is no labeled data from the target domain available during training, semi-supervised domain adaptation (SSDA) additionally allows models to access a small number of labeled target samples. Existing SSDA methods can be broadly categorized into cross-domain alignment, adversarial training, and semi-supervised learning-based approaches. Cross-domain alignment methods~\cite{li2021ecacl,yi2023source,singh2021improving} focus on reducing domain shifts by integrating complementary alignment techniques. Adversarial training-based methods~\cite{kim2020attract,qin2021contradictory,jiang2020bidirectional} leverage entropy minimization between class prototypes and neighboring unlabeled target samples to encourage domain-invariant representations. Semi-supervised learning-based approaches~\cite{yan2022multi,huang2023semi} utilize consistency regularization to enhance the model’s adaptation to the target domain with limited labeled data. While both our mixed-domain semi-supervised medical image segmentation (MiDSS) and SSDA tackle the challenges of limited annotation and domain shifts, the scenarios they address are fundamentally different. Specifically, SSDA assumes access to a substantial amount of labeled source data, a small set of labeled target samples, and unlabeled data solely from the target domain. In contrast, MiDSS operates under more challenging and practical conditions: the labeled data are limited and confined to a single source domain, while the unlabeled data come from both the source domain and multiple target domains. This unique data distribution increases the difficulty of effective domain adaptation compared to SSDA.}

\subsection{Data Augmentation via Copy-paste.} 
Copy-paste (CP)~\cite{yun2019cutmix} pastes the content of a specific region from one image onto the corresponding region of another, creating a new image that retains the semantic information from both original images. Compared to other pixel-level fusion strategies, such as MixUp~\cite{zhang2017mixup}, CP excels in preserving and blending semantic information from source images. While MixUp combines global source images proportionally, potentially leading to ambiguity issues when pixels from different classes are blended. Therefore, CP is a more suitable augmentation technique for medical image segmentation tasks. In line with this research direction, BCP~\cite{bai2023bidirectional} takes into account the dual embedding directions between unlabeled and labeled data, randomly selecting one of them to generate intermediate samples. 
To mitigate domain shift, previous works~\cite{fan2022ucc,bai2023bidirectional} employed CP to transfer knowledge from labeled data to unlabeled data by generating intermediate samples. Despite the progress, they fall short in harnessing information from intermediate domains, leading to sub-optimal transfer effects. In contrast, our method constructs comprehensive intermediate domains through semantic and stylistic mixing between labeled-unlabeled and unlabeled-unlabeled data at the data augmentation level. At the training strategy level, we leverage intermediate domain information to enhance model training on unlabeled data.

\section{Method}
\label{method}

Sec.~\ref{subsec::setting} introduces the MiDSS setting and defines relevant symbols. Sec.~\ref{subsec::method-UST} and Sec.~\ref{subsec:method-RUN} present UST and the extended framework UST-RUN. Sec.~\ref{subsec::loss} involves the overall loss functions, and Sec.~\ref{subsec::disscusion} provides further discussion.

\subsection{Notation and Problem Setting}
\label{subsec::setting}
In MiDSS scenario, the domain gap exists among images originating from $K$ data centers $\{\mathcal{D}_i\}_{i=1}^K$. The training set comprises $N$ labeled images $\{(x_i,y_i)\}_{i=1}^N$ from a single domain $\mathcal{D}_j \in \{\mathcal{D}_i\}_{i=1}^K$ and $M$ unlabeled images $\{u_i\}_{i=1}^M$ from multiple domains $\mathcal{D}_1,\ldots,\mathcal{D}_K$, where $M > N$. The image resolution of $H \times W \times D$ sequentially represents height and width and channel, and $y_i \in \{0,1\}^{H \times W \times C}$ is the ground truth of $x_i$, where $C$ is the number of class. We aim to train a model capable of delivering outstanding segmentation performance across all domains with or without labeled data. 

Based on the mean teacher framework~\cite{tarvainen2017mean}, teacher model $\hat{f}$ generates probability map $\hat{p}^w$ and pseudo-label $\hat{q}^w$ for weakly augmented unlabeled data $u^w$, guiding the student model $f$ in making predictions $p^s$ for strongly augmented unlabeled data $u^s$. $y^w$ supervises the student model $f$ on the predictions for weakly augmented labeled data $x^w$:
\begin{equation}
    \begin{split}
        x^w=\mathcal{A}^w(x)&; y^w=\mathcal{A}^w(y); u^w=\mathcal{A}^w(u); u^s=\mathcal{A}^s(u),\\
        \hat{p}^w=\hat{f}&(u^w); p^s=f(u^s); \hat{q}^w=\arg\max(\hat{p}^w),\\
    \end{split}
\end{equation}
where $\arg\max(\cdot)$ applied to probability distributions produces ``one-hot" probability distributions, and $\hat{f}$ is updated by the EMA of $f$~\cite{tarvainen2017mean}. The weak augmentation $\mathcal{A}^w$ includes cropping, rotation, flip, and elastic distortion, while the strong augmentation $\mathcal{A}^s$ builds upon $\mathcal{A}^w$ by incorporating non-geometric operations like color jitter and Gaussian blurring~\cite{sohn2020fixmatch,upretee2022fixmatchseg}. The overall framework of our method is shown in Fig.~\ref{framework}. The notations in framework are summarized in Tab.~\ref{tab:notations}.

\subsection{The UST Framework}
\label{subsec::method-UST}
\subsubsection{Intermediate Samples Generation by UCP}
\begin{figure}[!t]
\centering
\includegraphics[width=0.9\columnwidth]{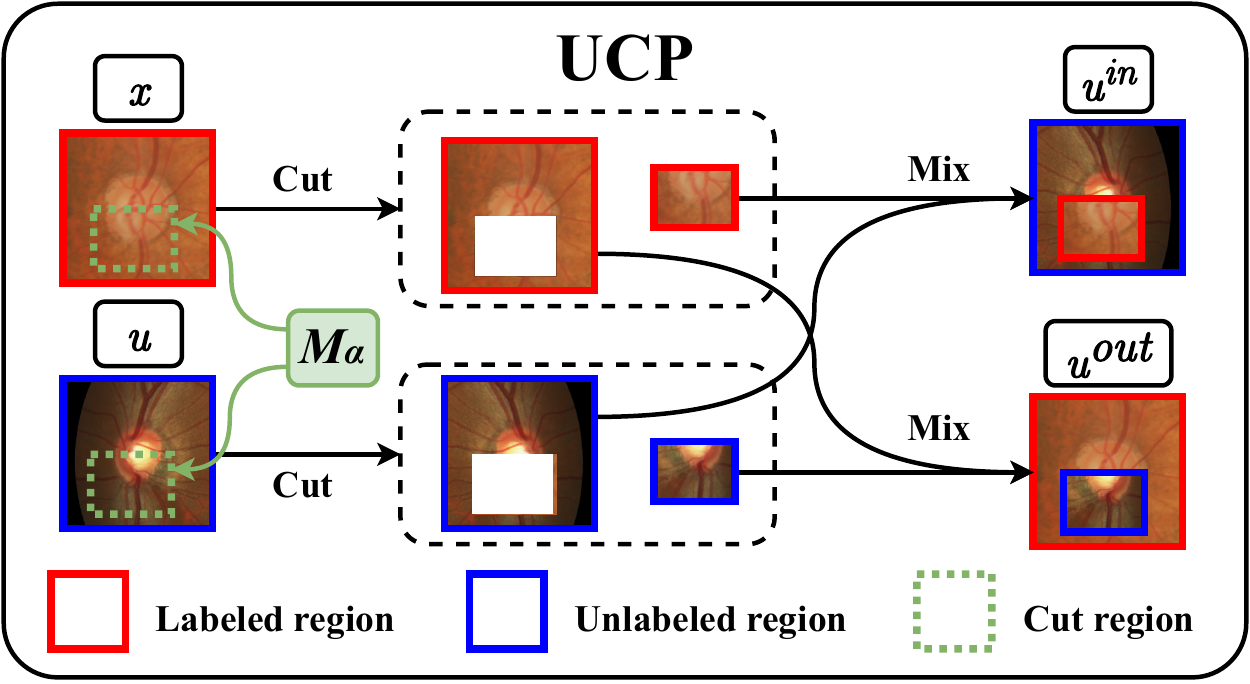}
\caption{\blue{The illustration of UCP between images. \textbf{Cut} refers to splitting the image into two parts according to $M_{\alpha}$, while \textbf{Mix} implies merging two parts of images back together.}}
\label{ucp}
\end{figure}

Given labeled data $(x^w,y^w)$ and unlabeled data $(u^s,\hat{p}^w,\hat{q}^w)$, we simultaneously apply both embedding directions of CP. This enables unified Copy-paste (UCP) between labeled and unlabeled data with distribution discrepancy. We obtain intermediate samples along with their pseudo-labels and probability maps:
\begin{equation}
    \label{CPformula}
    \begin{gathered}
        u^s_{in}=x^w \odot M_\alpha + u^s \odot (\mathbf{1}-M_\alpha),\\
        \hat{p}^w_{in}=y^w \odot M_\alpha + \hat{p}^w \odot (\mathbf{1}-M_\alpha),\\
        \hat{q}^w_{in}=y^w \odot M_\alpha + \hat{q}^w \odot (\mathbf{1}-M_\alpha),\\
        u^s_{out}=u^s \odot M_\alpha + x^w \odot (\mathbf{1}-M_\alpha),\\
        \hat{p}^w_{out}=\hat{p}^w \odot M_\alpha + y^w \odot (\mathbf{1}-M_\alpha),\\
        \hat{q}^w_{out}=\hat{q}^w \odot M_\alpha + y^w \odot (\mathbf{1}-M_\alpha),
    \end{gathered}
\end{equation}
where $M_\alpha\in\{0,1\}^{W \times H}$ is a randomly generated one-centered mask, indicating the region for CP. $\mathbf{1}$ indicates an all-one matrix, and $\odot$ means element-wise multiplication. \bluee{We follow the region selection strategy used in UniMatch~\cite{yang2023revisiting}. Specifically, a rectangular region is first generated by randomly sampling a target area and aspect ratio within predefined ranges. This region is then randomly positioned within the spatial dimensions of the image.} Fig.~\ref{ucp} illustrates the UCP between images, with the same operation applied to both probability maps and pseudo-labels.

\subsubsection{Symmetric Guided Training with UCP}
We aim to facilitate the knowledge transfer from labeled data to unlabeled data while maximizing the extraction of latent information from the unlabeled data. 
During training, we employ UCP between $(x^w, y^w)$ and $(u^s, \hat{p}^w, \hat{q}^w)$ to generate intermediate samples, probability maps and pseudo-labels by Eq.~\eqref{CPformula}. 
We set $w_i=\mathds{1}(\max (\hat{p}^w_i) \ge \tau)$ to indicate whether the pseudo-label is reliable, 
where $w_i$ is $i^{th}$ pixel of weight map $w$, and $\tau$ is a pre-defined confidence threshold. The indicator function is denoted as $\mathds{1}(\cdot)$. The pseudo-labels $\hat{q}^w_{in}$ and $\hat{q}^w_{out}$ guide the student model in predicting $p^{s}_{in}$ and $p^{s}_{out}$ for $u^{s}_{in}$ and $u^{s}_{out}$:
\begin{equation}
    \begin{gathered}
        \mathcal{L}_{in}=\mathcal{L}_{ce}(\hat{q}^w_{in}, p^{s}_{in},w_{in})+\mathcal{L}_{dice}(\hat{q}^w_{in}, p^{s}_{in}, w_{in}),\\
        \mathcal{L}_{out}=\mathcal{L}_{ce}(\hat{q}^w_{out}, p^{s}_{out}, w_{out})+\mathcal{L}_{dice}(\hat{q}^w_{out}, p^{s}_{out}, w_{out}),
    \end{gathered}
\end{equation}
where $\mathcal{L}_{ce}$ and $\mathcal{L}_{dice}$ respectively represent the cross-entropy loss and dice loss, which are formulated as:
\begin{equation}
    \begin{gathered}
        \mathcal{L}_{ce}(y,p,w)=-\frac{1}{H \times W}\sum_{i=1}^{H \times W} w_i y_i\log p_i,\\
        \mathcal{L}_{dice}(y,p,w)=1-\frac{2\times \sum_{i=1}^{H \times W}w_i p_i y_i}{\sum_{i=1}^{H \times W} w_i(p_i^2+y_i^2)},
    \end{gathered}
\end{equation}
where $p_i$, $y_i$ denote the probability of foreground and pseudo-label of the $i^{th}$ pixel, respectively.

\begin{figure}[!t]
\centering
\includegraphics[width=0.9\columnwidth]{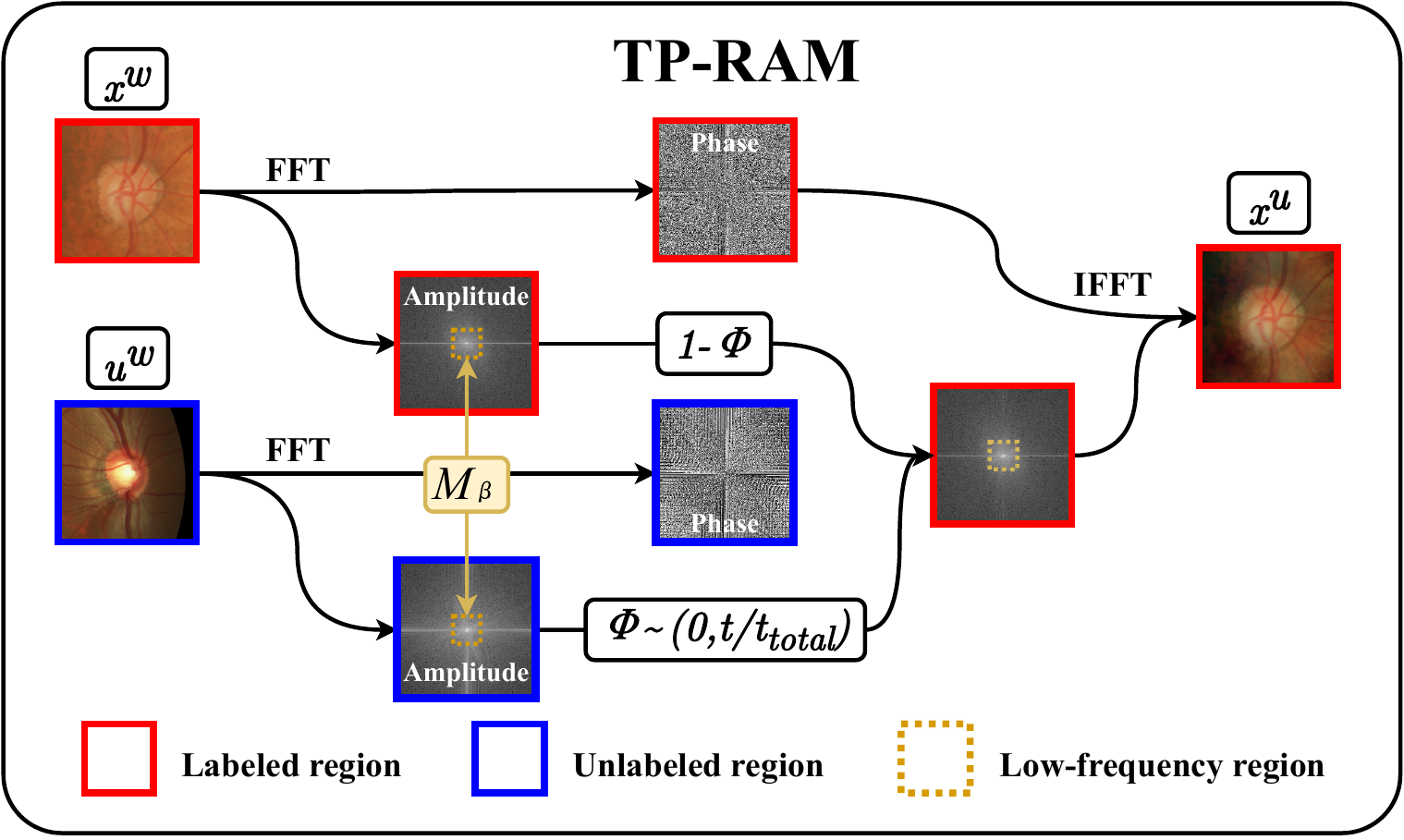}
\caption{\blue{The illustration of TP-RAM. Fast Fourier transform \textbf{(FFT)} extracts amplitude and phase maps of $x^w$ and $y^w$. The low-frequency components of two phase maps are mixed. $x^u$ is synthesized through inverse Fast Fourier transform \textbf{(IFFT)}.}}
\label{tpram}
\end{figure}

The supervision mentioned above effectively promotes the model training in intermediate domains, facilitating adaptation to unlabeled data. 
However, single-direction supervision lacks further utilization of intermediate domains information, resulting in sub-optimal performance in the domains of unlabeled data. Thus, we design a Symmetric Guidance training strategy (SymGD) to fully utilize the information from intermediate domains. In addition to $\mathcal{L}_{in}$ and $\mathcal{L}_{in}$, we also explore the guidance from intermediate samples to unlabeled data.

Similarly, we mix $x^w$ and $u^w$ to generate $u^{w}_{in}$ and $u^{w}_{out}$. The teacher model predicts pseudo-labels $\hat{q}^{ww}_{in}$ and $\hat{q}^{ww}_{out}$ on them. We merge the unlabeled regions of the pseudo-labels:
\begin{equation}
\begin{split}
    \label{pseudo_label_concat}
    \hat{q}^{mg} = \arg\max(\hat{f}(u_{out}^w)) \odot& M_\alpha + \\
    \arg&\max(\hat{f}(u_{in}^w)) \odot (\mathbf{1}-M_\alpha).
\end{split}
\end{equation}

We obtain the pseudo-label $\hat{p}^{mg}$ from intermediate samples and integrate it with $\hat{q}^w$ to provide better guidance for the prediction on $u^s$. The weight map $w^{ens}$ of $\hat{p}^{mg}$ is defined as:
\begin{equation}
    \label{ensemble}
    w^{ens}=(1 - (\hat{q}^w \otimes \hat{q}^{mg})) \odot w \odot w^{mg},
\end{equation}
where $\otimes$ is the pixel-wise XOR operator to indicate the consistency between pseudo-labels from two perspectives. $w$ and $w^{mg}$ represent the weight map of $\hat{q}^w$ and $\hat{q}^{mg}$, respectively. The loss of such direction can be defined as:
\begin{equation}
    \label{symmetric loss}
    \mathcal{L}_{sym} = \mathcal{L}_{ce}(\hat{q}^{mg},p^s,w^{ens})+\mathcal{L}_{dice}(\hat{q}^{mg},p^s,w^{ens}).
\end{equation}

To generate reliable pseudo-labels from intermediate samples, it is necessary that the model performs well in intermediate domains. Therefore, $\mathcal{L}_{sym}$ maintains a relatively low weight in the early training and rapidly increases later.

\subsubsection{Style Transition in Intermediate Samples}
UCP focuses on local semantic blending while overlooking the transition of stylistic differences between domains. Distribution information (\textit{i.e.}, style) is typically represented in low-frequency components, while edge information resides in high-frequency components~\cite{liu2021feddg}.
By interpolating low-frequency information, amplitude MixUp generates labeled data with a new style, which is then mix with $u_w$ to introduce style-transition components in intermediate domains.

However, arbitrary style transfer hinders stable construction of intermediate domains. To ensure the gradual stylistic transition of intermediate samples while retaining the diversity introduced by random MixUp, we propose Training Progress-aware Random Amplitude Mixup (TP-RAM).

As exhibited in Fig.~\ref{tpram}, we obtain the frequency space signals of image $x$ through Fast Fourier transformation $\mathcal{F}$ as follows:
\begin{equation}
    \label{fourier}
    \mathcal{F}(x)(u,v)=\sum_{h, w}x(h,w)e^{-j2 \pi (\frac{h}{H}u+\frac{w}{W}v)},
\end{equation}
where $j^2=-1$. Let $\mathcal{F}^{A}$, $\mathcal{F}^{P}$ be the amplitude and phase components of $\mathcal{F}$. For labeled and unlabeled data $(x^w,u^w)$, we obtain their amplitude $(\mathcal{F}^A(x^w), \mathcal{F}^A(u^w))$ and phase $(\mathcal{F}^P(x^w), \mathcal{F}^P(u^w))$. After that, we blend the low-frequency information of $\mathcal{F}^A(u^w)$ into $\mathcal{F}^A(x^w)$, and obtain a new image $x^u$ through inverse Fast Fourier transformation $\mathcal{F}^{-1}$:
\begin{equation}
    \label{xu}
    \begin{split}
        x^u = \mathcal{F}^{-1}[M_\beta \odot \mathcal{F}^A(u^w)& +\\ 
        (\mathbf{1}-M_\beta)\odot&\mathcal{F}^A(x^w),\mathcal{F}^P(x^w)],
    \end{split}
\end{equation}
where $M_{\beta}$ is a mask with values of $0$ except in the central $2\beta W \times 2\beta H$ region, and the values range from $0$ to $\Phi$, indicating the mixing ratio for low-frequency amplitude. 
$\Phi$ is defined as $\Phi(t)=\frac{t}{t_{total}}$, where $t$ denotes current training step and $t_{total}$ is the maximum number of steps.

\bluee{Style transfer is performed in a \textit{sample-to-sample} manner---that is, the style from one specific unlabeled sample is transferred directly to one labeled sample---rather than in a \textit{domain-to-sample} fashion, which would require explicit knowledge of domain identities or style clusters. Therefore, TP-RAM does not require distinguishing among styles of different target domains. By gradually increasing the intensity of style injection via TP-RAM, the labeled samples can effectively simulate a broader data distribution while still benefiting from precise ground truth supervision. This strategy enhances the model’s ability to generalize to multiple target domains.}

\begin{figure}[!t]
\centering
\includegraphics[width=\columnwidth]{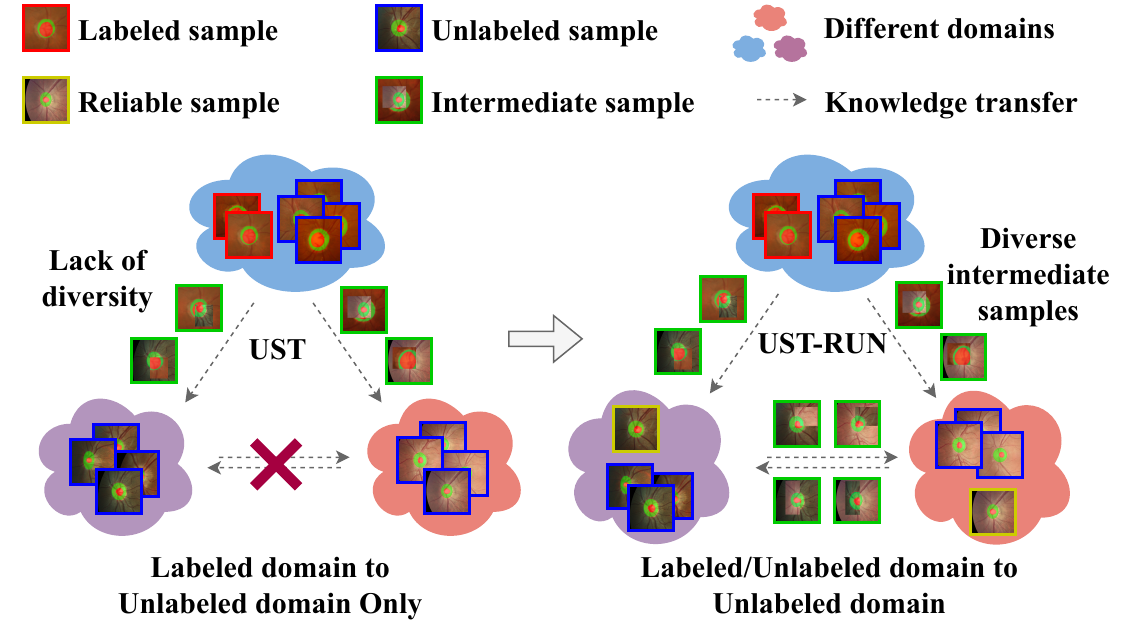}
\caption{\blue{In UST, intermediate samples are solely generated through copy-paste between labeled and unlabeled data, restricting the model’s training on unlabeled domains beyond the transferred knowledge from the labeled domain. In contrast, UST-RUN generates more diverse intermediate samples between any domains, enabling the model to leverage knowledge from easier-to-adapt unlabeled domains and enhance training across all domains.}}
\label{ust2run}
\end{figure}

\subsection{The UST-RUN Framework}
\label{subsec:method-RUN}
UST facilitates knowledge transfer from labeled data to other domains, but limited labeled data from a single domain results in poor diversity of intermediate samples. Moreover, low-quality pseudo-labels of intermediate samples from source images with poor pseudo-label quality can lead to negative transfer. To enhance diversity, we select reliable samples to generate intermediate samples. For efficient knowledge transfer to unreliable samples, we generate intermediate samples with high-quality pseudo-labels.

\subsubsection{Diverse Intermediate Samples Generation}
\label{subsubsec::R}
As shown in Fig.~\ref{ust2run}, we first aim to select reliable samples, namely unlabeled data with high-quality pseudo-labels.
Achieving consensus on a hard instance is more challenging for teacher and student models~\cite{grill2020bootstrap,Wang_Xu_Liu_Zhang_Fu_2020}. Thus, we estimate a hardness score $h$ for $u^w$ to indicate the difficulty to segment it. A smaller $h$ suggests a more reliable segmentation, and vice versa. 
Specifically, we obtain pseudo-label $\hat{q}^w$ from $\hat{f}$ and pseudo-label $q^w$ from $f$. By calculating the difference between them, we estimate the hardness score of $u^w$:
\begin{equation}
    h = 1 - \text{Dice}(\hat{q}^w, q^w),
\end{equation}
where $\text{Dice}(\cdot)$ is a commonly used metric to evaluate the similarity between segmentation maps, ranging from 0 to 1. The higher the dice score, the higher the similarity.

Then, we introduce the threshold $\gamma$ to compare with the hardness score for selecting reliable samples. The unlabeled sample with hardness less than $\gamma$ (\textit{i.e.}, $h < \gamma$) is considered as reliable, and is appended to a reliable samples queue $\mathcal{Q}$ with a maximum capacity of $K$. The corresponding pseudo-label, confidence map, and hardness score are also appended simultaneously. Notably, the filtering criterion is dynamic and adjusts throughout the training process to ensure high-quality filtering results, as the hardness values $h$ of all samples gradually decrease during training. In contrast, a fixed $\gamma$ cannot adapt to the evolving model predictions.
We design an adaptive dynamic threshold as follows.
Once the number of samples in $\mathcal{Q}$ exceeds $K$, the earliest sample that entered the queue is popped out, and $\gamma$ is updated to the maximum hardness among all the reliable samples in the current queue:
\begin{equation}
    \gamma = \max(h_{r,i}), i=1,2,\ldots,K,
\label{updateless}
\end{equation}
where $h_{r,i}$ represents the hardness score of the $i^{th}$ reliable sample in $Q$. The threshold continuously decreases since the model becomes confident, making the selection criterion increasingly stringent. However, to prevent the threshold from becoming overly strict, it is also periodically increased as needed.
An overly strict threshold results in almost no samples being selected as reliable. Therefore, if no samples are selected in one iteration, indicating that the selection criterion is too stringent, we slightly increase the threshold: 
\begin{equation}
    \gamma = \max(\gamma_0, \delta\gamma),
\label{updategreater}
\end{equation}
where $\gamma_0$ is the initial value of $\gamma$, and $\delta$ is a hyperparameter slightly greater than one, set to 1.0005 in our experiments.

For a triplet $(u^w_r, \hat{p}^w_r, \hat{q}^w_r)$—comprising a sample, confidence map, and pseudo-label randomly selected from $\mathcal{Q}$—and unlabeled data $u^s$, we generate more diverse intermediate samples by replacing the label with pseudo-label Eq.~\eqref{CPformula}, relying not only on labeled data for knowledge transfer.

\subsubsection{Enhanced Training for Unreliable Sample}
\label{subsubsec::Un}
Regarding the selection of unreliable samples, we do not require the same stringent criteria as for reliable samples. This distinction arises from the crucial role reliable samples play in guiding the training process. For unreliable samples, our emphasis is on developing a more effective training strategy which facilitates training on unlabeled data.
To achieve trade-off between effectiveness and computational costs, we select the sample with the highest hardness as unreliable sample in each batch.

To enhance knowledge transfer from labeled data to unreliable samples, the mask $M_{ur}$ is crafted by blending the predicted foreground region of the pseudo-label with the foreground region of the labeled data instead of random generation. The integration retains the background information of hard samples while preserving the precise foreground details from the labeled data. In this way, the training of intermediate samples benefits from the precise supervision, promoting effective knowledge transfer.

Specifically, for the unreliable sample $u^w_{ur}$ selected from the previous batch and the labeled data $x^w$, we merge the predicted foreground region of $\hat{q}^w_{ur}$ (pseudo-label for $u^w_{ur}$) with the foreground region of $y^w$, and take its bounding box region:
\begin{equation}
    M_{ur} = \text{Box}(\hat{q}^w_{ur} \oplus y^w),
\end{equation}
where $\oplus$ represents the pixel-wise OR operator, and \text{Box}$(\cdot)$ denotes bounding box region of foreground. To mitigate adverse effects of low-quality pseudo-labels, we only embed labeled data regions represented by $M_{ur}$ into unreliable samples without requiring reverse CP direction. The generated intermediate samples are then included in the calculation of $\mathcal{L}_{in}$.

\subsection{Loss Function}
\label{subsec::loss}
For each pair of labeled and unlabeled data, the overall objective function consists of a supervised loss, defined as: 
\begin{equation}
\label{eq:supervised_loss}
\mathcal{L}_{s}=\mathcal{L}_{ce}(y^w, f^{s}(x^w),\mathbf{1})+\mathcal{L}_{dice}(y^w, f^{s}(x^w), \mathbf{1}),
\end{equation} and unsupervised loss calculated through symmetric guidance:
\begin{equation}
    \label{total_loss}
    \mathcal{L}_{total}=\mathcal{L}_s+\lambda(\mathcal{L}_{in}+\mathcal{L}_{out}+\lambda\mathcal{L}_{sym}),
\end{equation}
where $\lambda$ is a weight coefficient decided by a time-dependent Gaussian warming-up function: $\lambda(t)=e^{-5(1-t/t_{total})}$. The factor $\lambda^2$ for $\mathcal{L}_{sym}$ increases relatively slowly in early training.

\subsection{Discussions}
\label{subsec::disscusion}
Following previous work~\cite{tarvainen2017mean}, our method consists of two models, both implemented based on U-Net~\cite{ronneberger2015u}. We promote domain knowledge transfer through intermediate samples generation and SymGD training strategy. Based on UST, we meticulously select reliable samples, generating more diverse intermediate samples with high-quality pseudo-labels. For unreliable samples with poor pseudo-label quality, we carefully retain their background information to generate intermediate samples and provide precise foreground guidance for them. 
During the testing phase, given a test image $x^{test}$, we obtain the probability map $p^{test}=f(x^{test})$ through the student model and subsequently derive the segmentation result. The inference parameters only involve those from the student model, resulting in efficiency comparable to U-Net~\cite{ronneberger2015u}.

\section{Experiments}
\label{experiments}

\subsection{Datasets}
\label{subsec::datasets}

\begin{table}[t]
  \centering
    \footnotesize
  \caption{Detailed partition information of four datasets. \# indicates the number of samples.}
\resizebox{0.7\linewidth}{!}{
  \begin{tabular}{lccc}
    \toprule
    \multirow{2}{*}{Dataset} & \multicolumn{2}{c}{Training} & \multirow{2}{*}{\#Testing} \\
    \cmidrule{2-3}
    & \#Labeled & \#Unlabeled &\\
    \midrule
    Fundus & 20 & 769 & 271 \\
    Prostate & 40 & 1,470 & 357 \\
    M\&Ms & 20 & 3,427 & 863\\
    BUSI & 64 & 454 & 129\\
    \bottomrule
  \end{tabular}}
  \label{partition}
  
\end{table}

\begin{figure}[t]
\centering
\includegraphics[width=\linewidth]{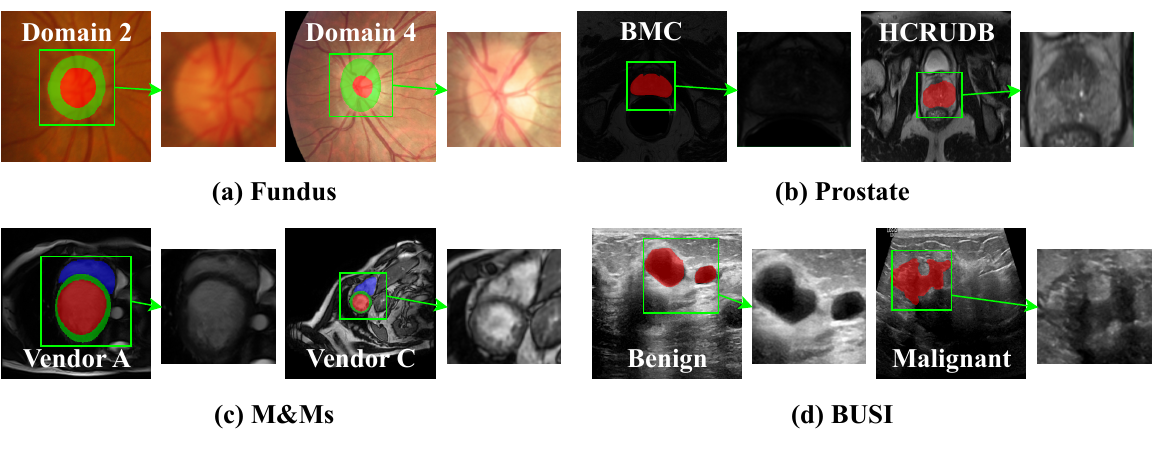}
\caption{Examples from the four datasets. Each pair of images within a dataset comes from different domains, exhibiting significant domain differences. Different colors in the images represent ground truths for different segmentation targets.} 
\label{dataset_examples}
\end{figure}

\begin{table*}[!t]
\centering
\footnotesize
\caption{\blue{Comparison of different methods on Fundus dataset. \#L represents the number of labeled samples. * denotes the Upper bound using all training samples in a domain as labeled data. $\uparrow$ indicates that a higher value corresponds to better performance, while $\downarrow$ suggests the opposite. The best performance is marked as \textbf{bold}, and the second-best is \underline{underlined}.}}
\resizebox{\linewidth}{!}{
\begin{tabular}{lccccccccc}
    \toprule
    \multirow{2}{*}{Method} & \multirow{2}{*}{\#L} & \multicolumn{4}{c}{DC $\uparrow$} & DC $\uparrow$ & JC $\uparrow$ & HD $\downarrow$ & ASD $\downarrow$\\
    \cmidrule{3-10}
    & & Domain 1 & Domain 2 & Domain 3 & Domain 4 & Avg. & Avg. & Avg. & Avg.\\
    \midrule
    U-Net & 20 & \blue{57.92\stdev{2.28} / 71.80\stdev{2.96}} & \blue{64.98\stdev{8.90} / 70.86\stdev{4.77}} & \blue{48.88\stdev{2.81} / 65.03\stdev{1.05}} & \blue{41.52\stdev{8.36} / 65.41\stdev{2.98}} & \blue{60.80\stdev{1.17}} & \blue{51.99\stdev{0.93}} & \blue{47.62\stdev{0.93}} & \blue{29.28\stdev{0.60}} \\
    UA-MT & 20 & \blue{63.53\stdev{5.91} / 79.63\stdev{1.65}} & \blue{67.60\stdev{6.39} / 79.69\stdev{7.40}} & \blue{45.32\stdev{14.26} / 56.69\stdev{12.67}} & \blue{41.63\stdev{7.71} / 60.45\stdev{7.11}} & \blue{61.82\stdev{7.89}} & \blue{52.31\stdev{7.50}} & \blue{42.47\stdev{8.73}} & \blue{25.79\stdev{7.86}} \\
    SIFA & 20 & \blue{59.89\stdev{13.04} / 75.98\stdev{0.96}} & \blue{66.76\stdev{3.28} / 84.47\stdev{5.34}} & \blue{64.47\stdev{3.95} / 85.31\stdev{2.18}} & \blue{57.24\stdev{3.07} / 73.84\stdev{4.48}} & \blue{70.99\stdev{4.54}} & \blue{58.48\stdev{5.25}} & \blue{18.59\stdev{2.22}} & \blue{11.76\stdev{1.17}} \\
    FixMatch & 20 & \blue{81.32\stdev{0.20} / 91.49\stdev{0.28}} & \blue{72.64\stdev{0.84} / 87.17\stdev{0.60}} & \blue{78.72\stdev{2.39} / 92.78\stdev{0.23}} & \blue{74.66\stdev{0.11} / 88.70\stdev{2.31}} & \blue{83.44\stdev{0.06}} & \blue{73.76\stdev{0.39}} & \blue{11.35\stdev{0.59}} & \blue{5.61\stdev{0.01}} \\
    UDA-VAE++ & 20 & \blue{60.75\stdev{8.11} / 79.64\stdev{1.59}} & \blue{69.16\stdev{0.41} / 86.38\stdev{0.62}} & \blue{67.17\stdev{5.57} / 85.16\stdev{0.35}} & \blue{68.63\stdev{0.30} / 79.56\stdev{1.87}} & \blue{74.56\stdev{1.48}} & \blue{62.31\stdev{1.29}} & \blue{17.42\stdev{0.26}} & \blue{9.62\stdev{0.34}} \\
    SS-Net & 20 & \blue{61.25\stdev{2.59} / 79.90\stdev{2.47}} & \blue{69.25\stdev{2.73} / 83.73\stdev{1.86}} & \blue{53.88\stdev{11.59} / 69.43\stdev{0.68}} & \blue{39.56\stdev{1.14} / 62.54\stdev{1.99}} & \blue{64.94\stdev{2.50}} & \blue{55.36\stdev{2.65}} & \blue{40.56\stdev{6.14}} & \blue{23.62\stdev{2.98}} \\
    BCP & 20 & \blue{75.29\stdev{5.15} / 90.63\stdev{0.66}} & \blue{78.92\stdev{2.45} / \textbf{91.73\stdev{0.37}}} & \blue{76.63\stdev{5.66} / 90.54\stdev{0.32}} & \blue{77.81\stdev{0.19} / 90.78\stdev{0.90}} & \blue{84.04\stdev{1.40}} & \blue{74.78\stdev{1.58}} & \blue{10.52\stdev{0.75}} & \blue{5.40\stdev{0.56}} \\
    CauSSL & 20 & \blue{66.53\stdev{4.45} / 83.38\stdev{3.93}} & \blue{67.28\stdev{0.35} / 81.66\stdev{1.34}} & \blue{55.36\stdev{8.24} / 71.80\stdev{11.19}} & \blue{39.36\stdev{0.09} / 48.04\stdev{1.97}} & \blue{64.17\stdev{3.34}} & \blue{54.15\stdev{3.32}} & \blue{36.44\stdev{6.80}} & \blue{21.85\stdev{2.96}} \\
    SAMed & 20 & \blue{62.18\stdev{0.27} / 91.53\stdev{0.30}} & \blue{74.97\stdev{0.08} / 90.41\stdev{0.15}} & \blue{70.19\stdev{0.25} / 90.24\stdev{0.18}} & \blue{62.16\stdev{0.23} / 81.51\stdev{0.69}} & \blue{77.90\stdev{0.21}} & \blue{67.96\stdev{0.08}} & \blue{13.00\stdev{0.05}} & \blue{7.29\stdev{0.09}} \\
    \blue{ABD} & \blue{20} & \blue{76.72\stdev{1.86} / 86.04\stdev{0.64}} & \blue{77.47\stdev{1.05} / 90.56\stdev{0.93}} & \blue{76.78\stdev{1.92} / 87.27\stdev{1.92}} & \blue{74.78\stdev{2.31} / 88.72\stdev{0.04}} & \blue{82.29\stdev{0.36}} & \blue{72.55\stdev{0.34}} & \blue{12.27\stdev{0.66}} & \blue{6.76\stdev{0.29}} \\
    \blue{H-SAM} & \blue{20} & \blue{76.78\stdev{0.27} / \underline{92.93\stdev{0.11}}} & \blue{78.97\stdev{0.05} / \underline{90.81\stdev{0.48}}} & \blue{76.47\stdev{0.54} / 91.80\stdev{0.09}} & \blue{79.85\stdev{1.67} / 92.65\stdev{0.32}} & \blue{85.03\stdev{0.24}} & \blue{76.12\stdev{0.19}} & \blue{9.88\stdev{0.30}} & \blue{5.16\stdev{0.24}} \\
    UST & 20 & \blue{\underline{83.16\stdev{0.78}} / 92.91\stdev{0.07}} & \blue{\underline{79.03\stdev{2.03}} / 89.49\stdev{0.62}} & \blue{\underline{83.28\stdev{1.27}} / \underline{92.82\stdev{0.21}}} & \blue{\underline{83.38\stdev{0.46}} / \underline{93.50\stdev{0.17}}} & \blue{\underline{87.20\stdev{0.66}}} & \blue{\underline{78.46\stdev{0.91}}} & \blue{\underline{8.09\stdev{0.18}}} & \blue{\underline{3.92\stdev{0.04}}} \\
    UST-RUN & 20 & \blue{\textbf{83.88\stdev{0.46}} / \textbf{93.34\stdev{0.22}}} & \blue{\textbf{80.44\stdev{0.77}} / 90.74\stdev{0.16}} & \blue{\textbf{84.49\stdev{0.41}} / \textbf{93.28\stdev{0.01}}} & \blue{\textbf{84.16\stdev{0.78}} / \textbf{93.72\stdev{0.04}}} & \blue{\textbf{88.01\stdev{0.11}}} & \blue{\textbf{79.51\stdev{0.43}}} & \blue{\textbf{7.70\stdev{0.00}}} & \blue{\textbf{3.71\stdev{0.07}}} \\
    \midrule
     UST &  10 & \blue{78.55\stdev{1.87} / 92.09\stdev{0.13}} & \blue{78.22\stdev{1.87} / 89.55\stdev{0.65}} & \blue{82.27\stdev{0.56} / 93.04\stdev{0.30}} & \blue{68.35\stdev{1.85} / 89.53\stdev{1.61}} & \blue{83.95\stdev{0.67}} & \blue{74.35\stdev{0.69}} & \blue{10.54\stdev{0.27}} & \blue{4.81\stdev{0.08}} \\
     UST-RUN &  10 & \blue{82.55\stdev{1.02} / 92.62\stdev{0.09}} & \blue{80.02\stdev{0.11} / 89.08\stdev{0.95}} & \blue{82.77\stdev{0.98} / 93.04\stdev{0.08}} & \blue{82.17\stdev{0.06} / 93.70\stdev{0.09}} & \blue{86.99\stdev{0.12}} & \blue{77.49\stdev{1.14}} & \blue{7.96\stdev{0.35}} & \blue{3.87\stdev{0.15}} \\
    \midrule
    \color{gray} Upper bound & \color{gray} * & \color{gray} \blue{85.47\stdev{0.08} / 93.38\stdev{0.04}} & \color{gray} \blue{80.60\stdev{0.07} / 90.94\stdev{0.05}} & \color{gray} \blue{85.28\stdev{0.22} / 93.02\stdev{0.04}} & \color{gray} \blue{85.65\stdev{0.06} / 93.24\stdev{0.04}} & \color{gray} \blue{88.45\stdev{0.02}} & \color{gray} \blue{80.32\stdev{0.04}} & \color{gray} \blue{7.42\stdev{0.01}} & \color{gray} \blue{3.73\stdev{0.04}} \\
    \bottomrule
\end{tabular}}
\label{fundus}
\end{table*}


\begin{figure}[!t]
\centering
\includegraphics[width=\linewidth]{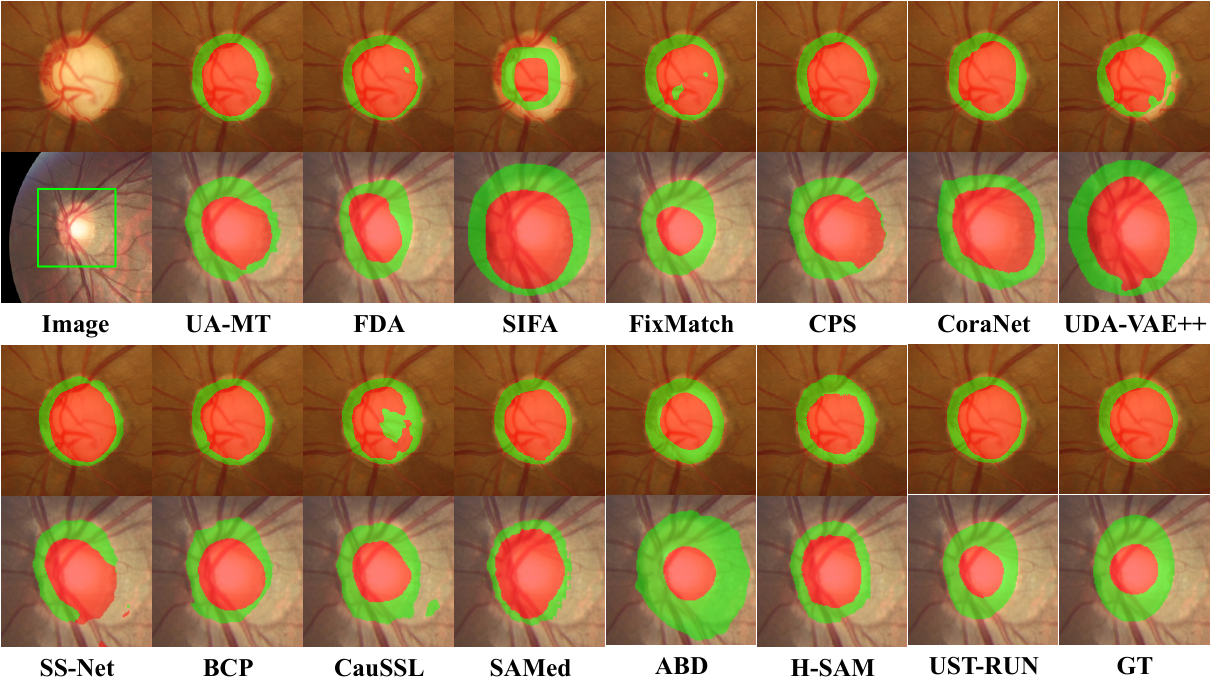}
\caption{\blue{Visual results from the Fundus dataset. The top test sample of each method is from the labeled domain (Domain 1), with the bottom from another domain (Domain 4). Red and green represent the Optical Cup and Disc, respectively.}}
\label{fundus_img}
\end{figure}
\begin{table*}[!t]
\centering
\footnotesize
\caption{\blue{Comparison of different methods on Prostate dataset.}}
\resizebox{\linewidth}{!}{
\begin{tabular}{lccccccccccc}
    \toprule
    \multirow{2}{*}{Method} & \multirow{2}{*}{\#L} & \multicolumn{6}{c}{DC $\uparrow$} & DC $\uparrow$ & JC $\uparrow$ & HD $\downarrow$ & ASD $\downarrow$\\
    \cmidrule{3-12}
    & & RUNMC & BMC & HCRUDB & UCL & BIDMC & HK & Avg. & Avg. & Avg. & Avg.\\
    \midrule
    U-Net & 40 & \blue{28.71\stdev{3.39}} & \blue{35.61\stdev{0.76}} & \blue{21.23\stdev{1.68}} & \blue{43.56\stdev{7.61}} & \blue{20.60\stdev{1.68}} & \blue{27.62\stdev{1.42}} & \blue{29.56\stdev{1.62}} & \blue{24.03\stdev{1.12}} & \blue{97.97\stdev{4.04}} & \blue{65.18\stdev{0.94}}\\
    UA-MT & 40 & \blue{27.41\stdev{2.86}} & \blue{16.66\stdev{16.94}} & \blue{9.91\stdev{3.64}} & \blue{40.40\stdev{1.39}} & \blue{15.57\stdev{3.35}} & \blue{14.65\stdev{5.04}} & \blue{20.77\stdev{0.57}} & \blue{15.00\stdev{0.17}} & \blue{121.23\stdev{12.95}} & \blue{79.17\stdev{2.26}}\\
    SIFA & 40 & \blue{72.66\stdev{0.01}} & \blue{70.28\stdev{0.13}} & \blue{62.94\stdev{1.62}} & \blue{67.97\stdev{7.81}} & \blue{69.12\stdev{3.54}} & \blue{64.99\stdev{0.24}} & \blue{67.99\stdev{2.23}} & \blue{54.86\stdev{2.72}} & \blue{37.03\stdev{10.75}} & \blue{14.21\stdev{1.67}}\\
    FixMatch & 40 & \blue{80.59\stdev{4.23}} & \blue{71.06\stdev{2.67}} & \blue{73.16\stdev{0.67}} & \blue{77.12\stdev{2.95}} & \blue{60.54\stdev{6.32}} & \blue{84.95\stdev{0.25}} & \blue{74.57\stdev{0.23}} & \blue{65.78\stdev{0.26}} & \blue{23.11\stdev{1.51}} & \blue{13.02\stdev{1.51}}\\
    UDA-VAE++ & 40 & \blue{67.62\stdev{1.58}} & \blue{67.70\stdev{2.34}} & \blue{52.02\stdev{19.06}} & \blue{65.82\stdev{1.94}} & \blue{66.19\stdev{4.09}} & \blue{65.37\stdev{0.30}} & \blue{64.12\stdev{3.42}} & \blue{50.67\stdev{3.01}} & \blue{38.17\stdev{5.61}} & \blue{18.04\stdev{3.62}}\\
    SS-Net & 40 & \blue{25.58\stdev{4.99}} & \blue{24.98\stdev{16.26}} & \blue{14.61\stdev{0.57}} & \blue{43.67\stdev{11.72}} & \blue{18.43\stdev{7.64}} & \blue{9.57\stdev{5.18}} & \blue{22.80\stdev{2.11}} & \blue{17.25\stdev{2.11}} & \blue{109.57\stdev{0.04}} & \blue{72.94\stdev{2.56}}\\
    BCP & 40 & \blue{65.39\stdev{6.73}} & \blue{73.27\stdev{1.83}} & \blue{47.41\stdev{1.77}} & \blue{63.64\stdev{6.67}} & \blue{74.75\stdev{0.76}} & \blue{62.62\stdev{6.86}} & \blue{64.51\stdev{0.42}} & \blue{54.65\stdev{0.73}} & \blue{51.07\stdev{2.16}} & \blue{25.46\stdev{2.49}}\\
    CauSSL & 40 & \blue{26.23\stdev{3.02}} & \blue{37.11\stdev{13.65}} & \blue{17.76\stdev{1.16}} & \blue{35.45\stdev{11.63}} & \blue{24.99\stdev{13.73}} & \blue{24.63\stdev{14.24}} & \blue{27.70\stdev{9.57}} & \blue{21.12\stdev{7.97}} & \blue{111.58\stdev{4.30}} & \blue{67.71\stdev{7.90}}\\
    SAMed & 40 & \blue{62.12\stdev{0.35}} & \blue{66.20\stdev{0.21}} & \blue{57.97\stdev{0.57}} & \blue{69.25\stdev{0.57}} & \blue{73.29\stdev{0.38}} & \blue{69.24\stdev{0.52}} & \blue{66.34\stdev{0.24}} & \blue{55.64\stdev{0.19}} & \blue{38.32\stdev{0.14}} & \blue{17.98\stdev{0.12}}\\
    \blue{ABD} & \blue{40} & \blue{60.94\stdev{3.38}} & \blue{68.70\stdev{4.08}} & \blue{13.98\stdev{3.60}} & \blue{58.22\stdev{1.07}} & \blue{72.07\stdev{1.45}} & \blue{24.77\stdev{5.69}} & \blue{49.78\stdev{0.11}} & \blue{40.44\stdev{0.05}} & \blue{64.46\stdev{5.64}} & \blue{37.38\stdev{2.45}}\\
    \blue{H-SAM} & \blue{40} & \blue{57.44\stdev{2.10}} & \blue{59.25\stdev{1.15}} & \blue{44.53\stdev{0.11}} & \blue{68.18\stdev{2.42}} & \blue{60.91\stdev{6.37}} & \blue{61.05\stdev{3.37}} & \blue{58.56\stdev{0.66}} & \blue{47.70\stdev{0.60}} & \blue{53.09\stdev{6.61}} & \blue{27.69\stdev{3.33}}\\
    UST & 40 & \blue{\underline{86.80\stdev{2.78}}} & \blue{\underline{86.34\stdev{0.01}}} & \blue{\underline{84.88\stdev{3.86}}} & \blue{\underline{86.69\stdev{2.33}}} & \blue{\textbf{88.36\stdev{0.37}}} & \blue{\underline{87.09\stdev{1.57}}} & \blue{\underline{86.69\stdev{1.82}}} & \blue{\underline{78.75\stdev{2.07}}} & \blue{\underline{11.83\stdev{2.07}}} & \blue{\underline{4.98\stdev{1.11}}}\\
    UST-RUN & 40 & \blue{\textbf{88.34\stdev{0.62}}} & \blue{\textbf{87.11\stdev{0.69}}} & \blue{\textbf{85.97\stdev{1.59}}} & \blue{\textbf{87.53\stdev{1.36}}} & \blue{\underline{88.13\stdev{0.64}}} & \blue{\textbf{88.00\stdev{0.70}}} & \blue{\textbf{87.51\stdev{0.93}}} & \blue{\textbf{79.47\stdev{1.38}}} & \blue{\textbf{11.39\stdev{1.25}}} & \blue{\textbf{4.80\stdev{0.78}}}\\
     \midrule
     UST &  20 & \blue{88.36\stdev{0.02}} & \blue{73.66\stdev{13.58}} & \blue{76.44\stdev{10.68}} & \blue{84.28\stdev{1.65}} & \blue{51.39\stdev{13.23}} & \blue{81.20\stdev{4.50}} & \blue{75.89\stdev{1.36}} & \blue{66.45\stdev{2.03}} & \blue{41.91\stdev{4.09}} & \blue{21.95\stdev{1.22}}\\
     UST-RUN &  20 & \blue{88.73\stdev{0.03}} & \blue{82.38\stdev{2.49}} & \blue{84.09\stdev{0.58}} & \blue{84.58\stdev{0.66}} & \blue{83.88\stdev{4.24}} & \blue{86.56\stdev{0.23}} & \blue{85.04\stdev{0.35}} & \blue{76.66\stdev{0.54}} & \blue{14.49\stdev{1.37}} & \blue{6.12\stdev{0.51}}\\
     \midrule
    \color{gray} Upper bound & \color{gray} * & \color{gray} \blue{88.56\stdev{0.06}} & \color{gray} \blue{88.56\stdev{0.08}} & \color{gray} \blue{85.60\stdev{0.15}} & \color{gray} \blue{88.63\stdev{0.03}} & \color{gray} \blue{88.90\stdev{0.12}} & \color{gray} \blue{89.44\stdev{0.06}} & \color{gray} \blue{88.28\stdev{0.05}} & \color{gray} \blue{80.67\stdev{0.05}} & \color{gray} \blue{10.07\stdev{0.04}} & \color{gray} \blue{4.13\stdev{0.02}}\\
    \bottomrule
\end{tabular}}
\label{prostate}
\end{table*}

\begin{figure}[!t]
\centering
\includegraphics[width=\linewidth]{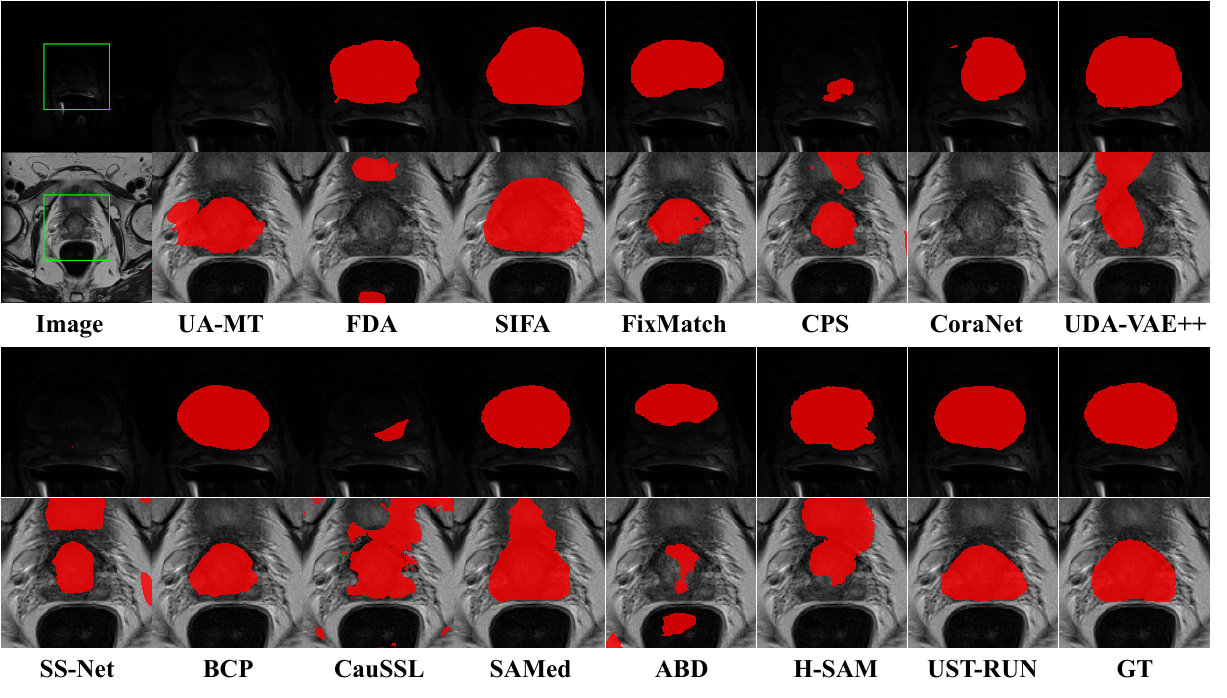}
\caption{\blue{Visual results of Prostate dataset. The test samples are from the labeled domain (BIDMC) and another domain (UCL). Red represents the location of the prostate.}}
\label{prostate_img}
\end{figure}
\begin{table*}[!t]
\setlength\tabcolsep{1.0mm}
\centering
\footnotesize
\caption{\blue{Comparison of different methods on M\&Ms dataset.}}
\resizebox{\linewidth}{!}{
\begin{tabular}{lccccccccc}
    \toprule
    \multirow{2}{*}{Method} & \multirow{2}{*}{\#L} & \multicolumn{4}{c}{DC $\uparrow$} & DC $\uparrow$ & JC $\uparrow$ & HD $\downarrow$ & ASD $\downarrow$\\
    \cmidrule{3-10}
    & & Vendor A & Vendor B & Vendor C & Vendor D & Avg. & Avg. & Avg. & Avg.\\
    \midrule
    U-Net & 20 & \blue{41.67\stdev{22.09}} / \blue{27.89\stdev{14.09}} / \blue{28.39\stdev{8.85}} & \blue{56.56\stdev{23.86}} / \blue{49.59\stdev{20.67}} / \blue{47.19\stdev{9.04}} & \blue{55.89\stdev{0.08}} / \blue{45.64\stdev{4.00}} / \blue{42.60\stdev{3.17}} & \blue{69.31\stdev{7.73}} / \blue{58.30\stdev{8.55}} / \blue{57.52\stdev{10.84}} & \blue{48.38\stdev{6.55}} & \blue{40.61\stdev{5.22}} & \blue{43.02\stdev{7.00}} & \blue{28.68\stdev{8.20}}\\
    UA-MT & 20 & \blue{40.48\stdev{3.47}} / \blue{30.44\stdev{6.97}} / \blue{20.91\stdev{8.44}} & \blue{50.65\stdev{15.84}} / \blue{43.24\stdev{15.60}} / \blue{36.96\stdev{14.67}} & \blue{48.55\stdev{7.66}} / \blue{37.35\stdev{2.39}} / \blue{30.98\stdev{3.46}} & \blue{47.27\stdev{7.61}} / \blue{43.11\stdev{6.88}} / \blue{34.01\stdev{11.17}} & \blue{38.66\stdev{0.99}} & \blue{30.14\stdev{1.42}} & \blue{71.75\stdev{0.85}} & \blue{41.14\stdev{0.43}}\\
    SIFA & 20 & \blue{59.95\stdev{5.69}} / \blue{34.39\stdev{0.36}} / \blue{38.16\stdev{1.29}} & \blue{60.77\stdev{8.05}} / \blue{39.03\stdev{8.40}} / \blue{37.38\stdev{8.25}} & \blue{49.58\stdev{9.81}} / \blue{37.69\stdev{2.74}} / \blue{36.92\stdev{2.30}} & \blue{58.05\stdev{2.14}} / \blue{39.13\stdev{2.12}} / \blue{44.25\stdev{5.65}} & \blue{44.61\stdev{3.58}} & \blue{32.25\stdev{3.83}} & \blue{24.44\stdev{0.80}} & \blue{10.78\stdev{0.95}}\\
    FixMatch & 20 & \blue{87.97\stdev{1.00}} / \blue{\underline{78.70\stdev{1.31}}} / \blue{78.95\stdev{2.56}} & \blue{90.32\stdev{1.05}} / \blue{81.61\stdev{1.65}} / \blue{80.80\stdev{2.45}} & \blue{88.25\stdev{0.58}} / \blue{80.37\stdev{0.42}} / \blue{75.56\stdev{3.49}} & \blue{90.42\stdev{0.62}} / \blue{81.20\stdev{0.78}} / \blue{81.48\stdev{0.50}} & \blue{82.97\stdev{0.02}} & \blue{74.02\stdev{0.04}} & \blue{6.49\stdev{0.39}} & \blue{3.45\stdev{0.09}}\\
    UDA-VAE++ & 20 & \blue{46.67\stdev{6.32}} / \blue{35.55\stdev{0.93}} / \blue{15.38\stdev{3.38}} & \blue{70.03\stdev{2.72}} / \blue{50.71\stdev{3.46}} / \blue{37.18\stdev{0.71}} & \blue{57.25\stdev{0.90}} / \blue{43.01\stdev{1.94}} / \blue{33.42\stdev{4.57}} & \blue{40.12\stdev{11.89}} / \blue{29.82\stdev{3.53}} / \blue{27.70\stdev{10.20}} & \blue{40.57\stdev{1.82}} & \blue{30.03\stdev{1.71}} & \blue{47.64\stdev{8.85}} & \blue{22.10\stdev{4.01}}\\
    SS-Net & 20 & \blue{40.10\stdev{12.55}} / \blue{25.92\stdev{1.55}} / \blue{21.34\stdev{0.98}} & \blue{49.29\stdev{14.68}} / \blue{44.00\stdev{12.28}} / \blue{39.09\stdev{2.50}} & \blue{58.61\stdev{5.22}} / \blue{48.42\stdev{9.93}} / \blue{45.58\stdev{8.43}} & \blue{51.87\stdev{4.48}} / \blue{48.06\stdev{5.66}} / \blue{36.84\stdev{1.36}} & \blue{42.43\stdev{2.32}} & \blue{34.88\stdev{1.43}} & \blue{53.40\stdev{5.51}} & \blue{36.52\stdev{5.62}}\\
    BCP & 20 & \blue{85.53\stdev{0.54}} / \blue{72.82\stdev{1.41}} / \blue{76.50\stdev{2.24}} & \blue{72.51\stdev{18.60}} / \blue{63.16\stdev{16.53}} / \blue{64.75\stdev{15.97}} & \blue{71.35\stdev{13.77}} / \blue{61.58\stdev{10.65}} / \blue{62.13\stdev{14.51}} & \blue{83.45\stdev{9.74}} / \blue{69.48\stdev{10.27}} / \blue{81.10\stdev{2.75}} & \blue{72.03\stdev{0.54}} & \blue{62.75\stdev{0.12}} & \blue{28.23\stdev{3.79}} & \blue{15.70\stdev{3.57}}\\
    CauSSL & 20 & \blue{43.06\stdev{4.04}} / \blue{28.50\stdev{9.29}} / \blue{16.39\stdev{8.39}} & \blue{47.72\stdev{4.62}} / \blue{38.45\stdev{5.95}} / \blue{30.91\stdev{1.46}} & \blue{48.55\stdev{10.60}} / \blue{38.76\stdev{6.73}} / \blue{30.49\stdev{0.76}} & \blue{52.30\stdev{2.10}} / \blue{44.14\stdev{8.80}} / \blue{34.88\stdev{6.27}} & \blue{37.84\stdev{3.40}} & \blue{28.82\stdev{2.95}} & \blue{75.07\stdev{3.07}} & \blue{38.74\stdev{1.06}}\\
    SAMed & 20 & \blue{72.23\stdev{0.38}} / \blue{52.50\stdev{1.76}} / \blue{35.16\stdev{0.21}} & \blue{79.68\stdev{0.36}} / \blue{64.91\stdev{0.16}} / \blue{54.23\stdev{0.28}} & \blue{81.34\stdev{0.31}} / \blue{63.88\stdev{0.01}} / \blue{56.80\stdev{0.46}} & \blue{79.44\stdev{0.06}} / \blue{71.68\stdev{0.92}} / \blue{55.14\stdev{0.23}} & \blue{63.92\stdev{0.15}} & \blue{53.23\stdev{0.28}} & \blue{18.16\stdev{0.33}} & \blue{8.83\stdev{0.23}}\\
    \blue{ABD} & \blue{20} & \blue{43.08\stdev{7.78}} / \blue{36.34\stdev{0.26}} / \blue{28.46\stdev{0.28}} & \blue{46.57\stdev{0.31}} / \blue{28.19\stdev{9.84}} / \blue{28.62\stdev{1.20}} & \blue{63.78\stdev{2.98}} / \blue{43.19\stdev{0.98}} / \blue{47.49\stdev{6.69}} & \blue{66.59\stdev{5.77}} / \blue{53.55\stdev{4.12}} / \blue{43.67\stdev{6.57}} & \blue{44.13\stdev{0.47}} & \blue{35.67\stdev{0.73}} & \blue{51.84\stdev{0.58}} & \blue{35.69\stdev{1.16}}\\
    \blue{H-SAM} & \blue{20} & \blue{60.17\stdev{0.76}} / \blue{40.99\stdev{2.29}} / \blue{42.25\stdev{2.42}} & \blue{67.20\stdev{7.45}} / \blue{52.19\stdev{3.50}} / \blue{53.19\stdev{3.25}} & \blue{73.17\stdev{2.94}} / \blue{53.11\stdev{0.88}} / \blue{52.92\stdev{1.12}} & \blue{73.93\stdev{1.58}} / \blue{55.00\stdev{2.26}} / \blue{55.64\stdev{3.23}} & \blue{56.65\stdev{1.28}} & \blue{46.10\stdev{0.81}} & \blue{29.98\stdev{1.62}} & \blue{17.25\stdev{1.73}}\\
    UST & 20 & \blue{\underline{88.36\stdev{0.83}}} / \blue{78.28\stdev{2.71}} / \blue{\underline{81.29\stdev{0.91}}} & \blue{\underline{90.36\stdev{1.58}}} / \blue{\underline{81.78\stdev{2.69}}} / \blue{\underline{82.58\stdev{1.58}}} & \blue{\textbf{90.13\stdev{1.24}}} / \blue{\underline{82.75\stdev{0.13}}} / \blue{\underline{82.19\stdev{0.11}}} & \blue{\underline{91.59\stdev{0.95}}} / \blue{\underline{82.47\stdev{0.18}}} / \blue{\underline{84.15\stdev{1.82}}} & \blue{\underline{84.66\stdev{0.50}}} & \blue{\underline{75.69\stdev{0.73}}} & \blue{\underline{4.89\stdev{0.37}}} & \blue{\underline{2.37\stdev{0.07}}}\\
    UST-RUN & 20 & \blue{\textbf{88.43\stdev{1.44}}} / \blue{\textbf{79.80\stdev{1.07}}} / \blue{\textbf{81.50\stdev{1.94}}} & \blue{\textbf{90.77\stdev{1.05}}} / \blue{\textbf{82.89\stdev{1.27}}} / \blue{\textbf{83.45\stdev{2.43}}} & \blue{\underline{89.94\stdev{1.75}}} / \blue{\textbf{83.00\stdev{0.57}}} / \blue{\textbf{82.94\stdev{0.26}}} & \blue{\textbf{91.98\stdev{0.40}}} / \blue{\textbf{83.21\stdev{0.03}}} / \blue{\textbf{84.69\stdev{1.27}}} & \blue{\textbf{85.22\stdev{0.73}}} & \blue{\textbf{76.25\stdev{0.90}}} & \blue{\textbf{4.56\stdev{0.28}}} & \blue{\textbf{2.14\stdev{0.18}}} \\
     \midrule
     UST & 5 & \blue{64.00\stdev{2.34}} / \blue{74.02\stdev{4.93}} / \blue{72.90\stdev{6.71}} & \blue{88.82\stdev{0.59}} / \blue{80.23\stdev{1.58}} / \blue{82.98\stdev{2.28}} & \blue{83.85\stdev{2.40}} / \blue{72.09\stdev{8.80}} / \blue{71.53\stdev{8.29}} & \blue{86.80\stdev{5.15}} / \blue{76.69\stdev{7.66}} / \blue{69.39\stdev{16.33}} & \blue{77.78\stdev{1.30}} & \blue{67.40\stdev{1.46}} & \blue{13.35\stdev{1.04}} & \blue{5.26\stdev{1.40}}\\
     UST-RUN &  5 &  \blue{74.41\stdev{2.22}} / \blue{75.77\stdev{2.57}} / \blue{75.40\stdev{4.01}} & \blue{89.31\stdev{0.32}} / \blue{80.62\stdev{1.65}} / \blue{83.40\stdev{1.70}} & \blue{88.44\stdev{2.86}} / \blue{79.34\stdev{0.37}} / \blue{77.64\stdev{4.09}} & \blue{89.38\stdev{1.96}} / \blue{83.14\stdev{1.03}} / \blue{80.15\stdev{3.80}} & \blue{81.42\stdev{0.24}} & \blue{71.62\stdev{0.33}} & \blue{7.53\stdev{1.58}} & \blue{3.76\stdev{0.86}}\\
     \midrule
    \color{gray} Upper bound & \color{gray} * & \color{gray} \blue{91.19\stdev{0.25}} / \blue{83.65\stdev{0.06}} / \blue{84.00\stdev{0.18}} & \color{gray} \blue{92.48\stdev{0.22}} / \blue{85.56\stdev{0.62}} / \blue{84.54\stdev{0.16}} & \color{gray} \blue{92.26\stdev{0.08}} / \blue{84.35\stdev{0.11}} / \blue{84.27\stdev{0.41}} & \color{gray} \blue{91.82\stdev{0.37}} / \blue{82.90\stdev{1.25}} / \blue{83.50\stdev{0.39}} & \blue{86.71\stdev{0.15}} & \color{gray} \blue{78.52\stdev{0.17}} & \color{gray} \blue{4.41\stdev{0.13}} & \color{gray} \blue{2.24\stdev{0.13}}\\
    \bottomrule
\end{tabular}
}
\label{mnms}
\end{table*}

\begin{figure}[!t]
\centering
\includegraphics[width=\linewidth]{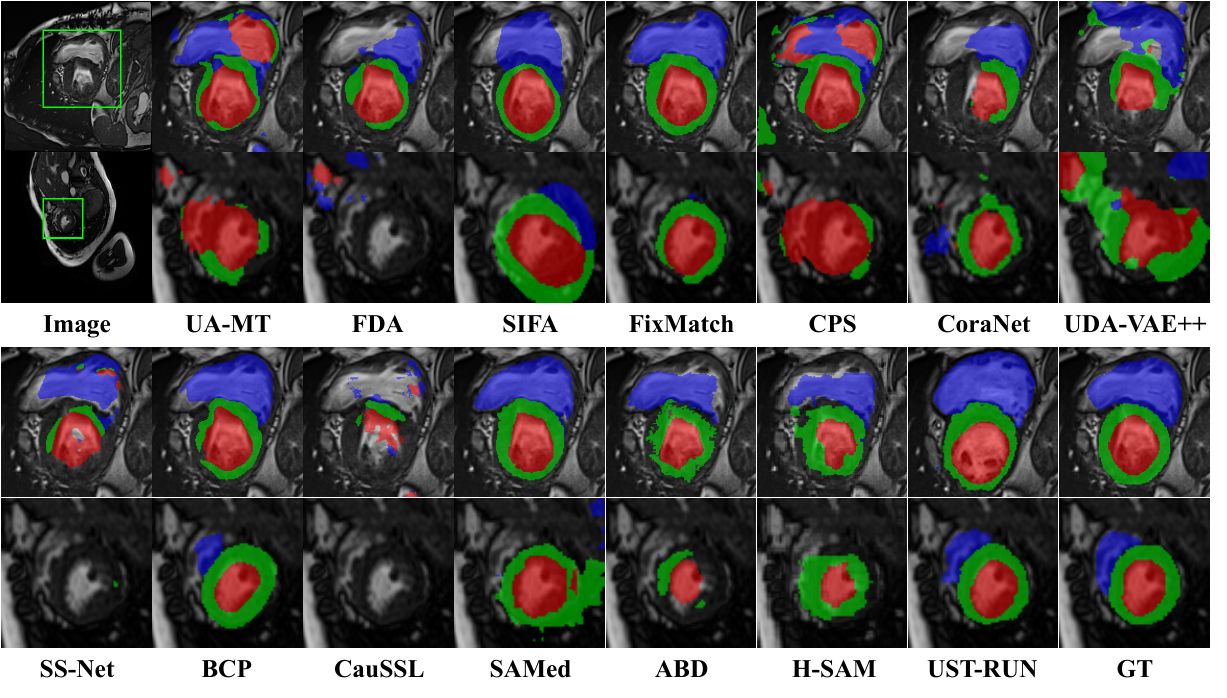}
\caption{\blue{Visual results of M\&Ms dataset. The first row shows results for a test sample from the labeled domain (Vendor A), and the second row displays results for the sample from another domain (Vendor D). Red, green, and blue represent LV, MYO, and RV, respectively.}}
\label{mnms_img}
\end{figure}
\begin{table}[!t]
\centering
\footnotesize
\caption{\blue{Comparison of different methods on BUSI dataset.}}
\resizebox{\linewidth}{!}{
\begin{tabular}{lccccccc}
    \toprule
    \multirow{2}{*}{Method} & \multirow{2}{*}{\#L} & \multicolumn{2}{c}{DC $\uparrow$} & DC $\uparrow$ & JC $\uparrow$ & HD $\downarrow$ & ASD $\downarrow$\\
    \cmidrule{3-8}
    & & Bengin & Malignant & Avg. & Avg. & Avg. & Avg.\\
    \midrule
    U-Net & 129 & \blue{59.67\stdev{0.15}} & \blue{65.15\stdev{1.27}} & \blue{62.42\stdev{0.56}} & \blue{51.48\stdev{1.17}} & \blue{54.06\stdev{0.28}} & \blue{22.91\stdev{0.53}}\\
    SIFA & 129 & \blue{59.55\stdev{6.58}} & \blue{55.00\stdev{0.42}} & \blue{57.28\stdev{3.08}} & \blue{45.39\stdev{2.21}} & \blue{50.63\stdev{0.78}} & \blue{21.23\stdev{0.19}}\\
    FixMatch & 129 & \blue{61.53\stdev{0.36}} & \blue{71.47\stdev{0.47}} & \blue{66.50\stdev{0.42}} & \blue{56.16\stdev{0.55}} & \blue{48.41\stdev{1.86}} & \blue{20.27\stdev{1.43}}\\
    UDA-VAE++ & 129 & \blue{64.43\stdev{1.50}} & \blue{62.42\stdev{1.34}} & \blue{63.43\stdev{1.41}} & \blue{50.36\stdev{1.76}} & \blue{48.26\stdev{0.33}} & \blue{19.44\stdev{0.35}}\\
    BCP & 129 & \blue{63.19\stdev{1.74}} & \blue{68.42\stdev{1.72}} & \blue{65.81\stdev{1.73}} & \blue{54.98\stdev{1.85}} & \blue{48.45\stdev{10.43}} & \blue{19.70\stdev{4.49}}\\
    \blue{ABD} & \blue{129} & \blue{54.77\stdev{6.84}} & \blue{65.67\stdev{1.23}} & \blue{60.22\stdev{4.04}} & \blue{49.61\stdev{3.34}} & \blue{47.12\stdev{1.46}} & \blue{20.32\stdev{2.70}}\\
    \blue{H-SAM} & \blue{129} & \blue{\textbf{68.74\stdev{3.00}}} & \blue{64.96\stdev{0.83}} & \blue{66.85\stdev{1.92}} & \blue{55.70\stdev{2.20}} & \blue{\underline{42.66\stdev{5.42}}} & \blue{19.80\stdev{2.37}}\\
    UST & 129 & \blue{62.87\stdev{1.68}} & \blue{\underline{71.72\stdev{0.53}}} & \blue{\underline{67.30\stdev{0.57}}} & \blue{\underline{57.02\stdev{0.32}}} & \blue{42.97\stdev{1.77}} & \blue{\underline{16.96\stdev{1.21}}}\\
    UST-RUN & 129 & \blue{\underline{66.03\stdev{2.11}}} & \blue{\textbf{73.04\stdev{0.82}}} & \blue{\textbf{69.53\stdev{0.64}}} & \blue{\textbf{59.53\stdev{0.86}}} & \blue{\textbf{41.50\stdev{2.16}}} & \blue{\textbf{16.37\stdev{1.58}}} \\
    \midrule
    UST & 64 & \blue{60.30\stdev{0.37}} & \blue{72.53\stdev{0.35}} & \blue{66.42\stdev{0.01}} & \blue{56.42\stdev{0.05}} & \blue{40.09\stdev{0.23}} & \blue{18.04\stdev{0.23}}\\
    UST-RUN & 64 & \blue{65.16\stdev{2.53}} & \blue{72.75\stdev{0.78}} & \blue{68.96\stdev{0.88}} & \blue{58.86\stdev{1.11}} & \blue{38.45\stdev{0.43}} & \blue{15.70\stdev{0.51}}\\
    \midrule
    \color{gray} Upper bound & \color{gray} * & \color{gray} \blue{71.73\stdev{0.14}} & \color{gray} \blue{72.70\stdev{0.36}} & \color{gray} \blue{72.22\stdev{0.25}} & \color{gray} \blue{61.52\stdev{0.19}} & \color{gray} \blue{38.15\stdev{0.08}} & \color{gray} \blue{14.95\stdev{0.08}}\\
    \bottomrule
\end{tabular}}
\label{busi}
\end{table}

\begin{figure}[!t]
\centering
\includegraphics[width=\linewidth]{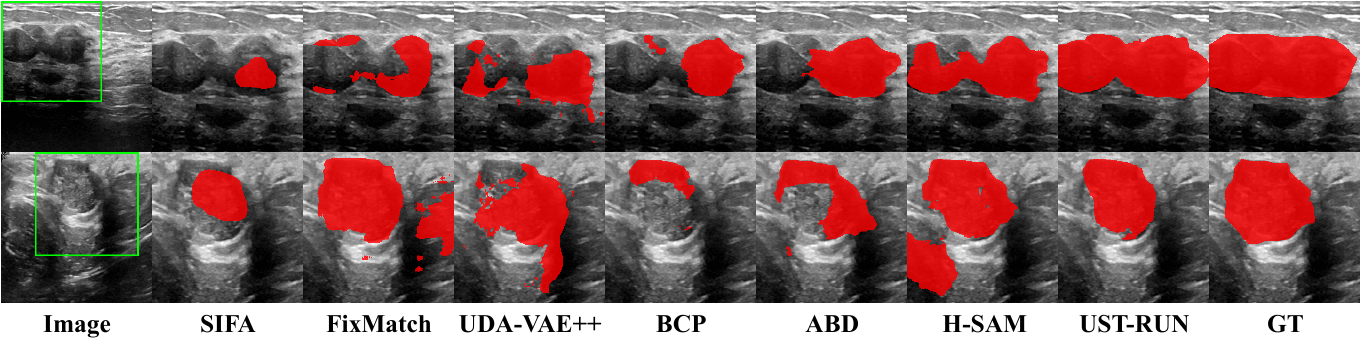}
\caption{\blue{Visual results of BUSI dataset. The first row shows results for a test sample from the labeled domain (bengin), and the second row displays results for one from another domain (malignant). Red represents the location of the breast tumor.}}
\label{busi_img}
\end{figure}
\begin{table}[t]
\centering
\footnotesize
\caption{Integration methods on Fundus dataset.}
\resizebox{0.7\linewidth}{!}{
\begin{tabular}{lccccc}
    \toprule
    \multirow{2}{*}{Method} & \multirow{2}{*}{\#L} & DC $\uparrow$ & JC $\uparrow$ & HD $\downarrow$ & ASD $\downarrow$\\
    \cmidrule{3-6}
    & & \multicolumn{4}{c}{Avg.}\\
    \midrule
    FixMatch & 20 & 83.39 & 73.48 & 11.77 & 5.60\\
    +FDA & 20 & 84.72 & 75.33 & 10.38 & 4.82\\
    +CutMix & 20 & 85.59 & 76.32 & 9.61 & 4.71\\
    +ClassMix & 20 & 84.35 & 75.02 & 10.84 & 5.59\\
    +CowMix & 20 & \underline{85.99} & \underline{77.07} & 9.28 & 4.56\\
    +FMix & 20 & 85.95 & 76.80 & \underline{9.26} & \underline{4.52}\\
    \midrule
    UST-RUN & 20 & \textbf{88.09} & \textbf{79.81} & \textbf{7.70} & \textbf{3.66}\\
    \bottomrule
\end{tabular}}
\label{fundus_ssmsuda}
\end{table}

\textbf{Fundus dataset}~\cite{wang2020dofe} comprises retinal fundus images collected from four medical centers, primarily for optic cup and disc segmentation tasks. Each image has been cropped to form a region of interest with a $800 \times 800$ bounding-box. We resize and randomly crop images to $256 \times 256$.

\textbf{Prostate dataset}~\cite{liu2020shape} contains prostate T2-weighted MRI data collected from six different data sources, with disparities in in-/through-plane resolution. We randomly split the dataset into training and testing sets based on a ratio of $4:1$, with each 2D slice resized and randomly cropped to $384 \times 384$.

\textbf{M\&Ms dataset}M\&Ms~\cite{campello2021multi} comprises cardiac images categorized into four distinct domains based on scanner manufacturers, primarily for the left ventricle (LV), left ventricle myocardium (MYO), and right ventricle (RV) segmentation tasks. We divide the data following the same criteria as Prostate dataset. Each slice is resized to $288 \times 288$.

\textbf{BUSI dataset}~\cite{al2020dataset} consists of 133 normal, 437 benign, and 210 malignant breast cancer ultrasound images. We only use the 647 abnormal images with breast cancer segmentation targets. We treat benign and malignant cases as two distinct domains. The dataset is split following the same criteria as Prostate dataset. Each image is resized to $256 \times 256$.

Examples from the four datasets are shown in Fig.~\ref{dataset_examples}. 
The detailed description of the datasets is shown in Tab.~\ref{partition}. 

\subsection{Implementation Details}
\label{subsec::details}
We implement our experiments on an NVIDIA GeForce RTX 3090 GPU. We set $\beta=0.01$, $\tau=0.95$ as default, and use Stochastic Gradient Descent (SGD) with a momentum of 0.9 and weight decay of 0.0001, with an initial learning rate of 0.03. The batch size is set as 8, including 4 labeled data and 4 unlabeled data. The iteration is set to 30,000 for Fundus and BUSI dataset, and 60,000 for Postate and M\&Ms dataset. We compare our method with other state-of-the-art (SOTA) methods, including SSMIS methods such as UA-MT~\cite{yu2019uncertainty}, FixMatch~\cite{sohn2020fixmatch}, SS-Net~\cite{wu2022exploring}, BCP~\cite{bai2023bidirectional}, and CauSSL~\cite{miao2023caussl} as well as UDA methods like SIFA~\cite{chen2020unsupervised} and UDA-VAE++~\cite{lu2022unsupervised}. Additionally, we compare with fine-tuned foundation model SAM~\cite{kirillov2023segment} by SAMed~\cite{zhang2023customized}.
\blue{It is important to note that although the vanilla FixMatch method was originally designed for classification tasks, its core idea can be extended to segmentation tasks by applying the same consistency regularization on a pixel-wise level. Specifically, the model generates pixel-level pseudo-labels from weakly augmented unlabeled images, and high-confidence pixels are used as supervision signals when strongly augmented images are input. This adaptation has been demonstrated to be effective in existing literature for semi-supervised medical image segmentation~\cite{upretee2022fixmatchseg}.}
In each experiment, a small amount of data from one domain is labeled, 
while the remainder serves as unlabeled data. 
In term of the upper bound, we employ UCP in FixMatch, using all available training data from a certain domain as labeled data to ensure sufficient source domain information, while also including unlabeled data from other domains. \blue{Compared to the standard MiDSS setting, the upper bound setting leverages a larger number of labeled data to provide more domain information from the labeled data.}
We adopt the Dice coefficient (DC), Jaccard coefficient (JC), 95\% Hausdorff Distance (HD), and Average Surface Distance (ASD) as evaluation metrics. 

\subsection{Comparison with State-of-the-Art Methods}
\label{subsec::comparison}
\subsubsection{Results on Fundus dataset}

\begin{table*}[t]
\centering
\footnotesize
\caption{Ablation experiments on Fundus dataset. The investigated settings are introduced as follows: 1) UCP: Generating intermediate samples by UCP, 2) VanillaGD: Guiding $p^s$ with $\hat{p}$, 3) SymGD: Guiding $p^s$ with mergence of $\hat{p}_{out}$ and $\hat{p}_{in}$, 4) TP-RAM: Gradually increasing the range of $\beta$, 5) RAM: $\beta$ is uniformly sampled from $0$ to $1$, 6) R: selecting reliable samples to generate more diverse intermediate samples, 7) Un: enhancing the model training on unreliable samples.}
\resizebox{0.8\linewidth}{!}{
\begin{tabular}{ccccccccccccc}
    \toprule
    \multicolumn{8}{c}{Task} & \multicolumn{5}{c}{Optic Cup / Disc Segmentation}\\
    \midrule
    \multirow{2}{*}{Method} & \multirow{2}{*}{UCP} & \multirow{2}{*}{VanillaGD} & \multirow{2}{*}{SymGD} & \multirow{2}{*}{TP-RAM} & \multirow{2}{*}{RAM} & \multirow{2}{*}{R} & \multirow{2}{*}{Un} & \multicolumn{5}{c}{DC $\uparrow$}\\
    \cmidrule{9-13}
    & & & & & & & & Domain 1 & Domain 2 & Domain 3 & Domain 4 & Avg.\\
    \midrule
    \#1 & \checkmark & & & & & & & 82.67 / 93.11 & 72.08 / 88.86 & 82.97 / 92.78 & 80.84 / 92.94 & 85.78\\
    \#2 & \checkmark & \checkmark & & & & & & 83.19 / 93.45 & 73.57 / 89.48 & 82.16 / 92.91 & 80.42 / 93.35 & 86.07\\
    \#3 & \checkmark & & \checkmark & & & & & 83.21 / 93.48 & 76.13 / 89.06 & 83.04 / 92.87 & 83.63 / 93.69 & 86.89\\
    \#4 & \checkmark & & & \checkmark & & & & 83.41 / \textbf{93.54} & 77.18 / 88.96 & 82.69 / 92.88 & 83.54 / 93.39 & 86.95\\
    \#5 & \checkmark & & \checkmark & & \checkmark & & & 83.27 / 93.41 & 76.14 / 88.46 & 83.27 / 92.90 & 83.64 / 93.40 & 86.82\\
    \#6 & \checkmark & & \checkmark & \checkmark & & & & 83.71 / 92.96 & \textbf{80.47} / 89.93 & 84.18 / 92.97 & 83.71 / 93.38 & 87.66\\
    \#7 & \checkmark & & \checkmark & \checkmark & & \checkmark & & \textbf{84.32} / 93.14 & 79.89 / 90.37 & 84.30 / 93.24 & 84.46 / 93.59 & 87.91\\
    \midrule
    \#8 & \checkmark & & \checkmark & \checkmark & & \checkmark & \checkmark & 84.21 / 93.50 & 79.90 / \textbf{90.63} & \textbf{84.78} / \textbf{93.27} & \textbf{84.72} / \textbf{93.70} & \textbf{88.09}\\
    \bottomrule
\end{tabular}}
\label{fundusablation}
\end{table*}
\begin{table}[!t]
\centering
\footnotesize
\caption{Varying the low-frequency size parameter $\beta$ and the confidence threshold $\tau$ on the Fundus dataset.}
\resizebox{0.7\linewidth}{!}{
\begin{tabular}{cccccc}
    \toprule
    $\beta$ & 0.2 & 0.1 & 0.05 & 0.01 & 0.005\\
    \midrule
    Avg. DC $\uparrow$ & 87.51 & 87.80 & 88.05 & \textbf{88.09} & 87.90\\
    \bottomrule
    \toprule
    $\tau$ & 0.85 & 0.90 & 0.95 & 0.98 & 0.99\\
    \midrule
    Avg. DC $\uparrow$ & 87.87 & 88.01 & \textbf{88.09} & 87.87 & 87.70\\
    \bottomrule
\end{tabular}}
\label{beta_tau}
\end{table}

\begin{table}[!t]
\centering
\footnotesize
\caption{Varying the increasing coefficient $\delta$ and the Queue capacity $K$ on the Prostate dataset.}
\resizebox{0.7\linewidth}{!}{
\begin{tabular}{cccccc}
    \toprule
    $\delta$ & 1.0001 & 1.0005 & 1.001 & 1.005 & 1.01\\
    \midrule
    Avg. DC $\uparrow$ & 87.90 & \textbf{88.17} & 88.04 & 87.29 & 85.89\\
    \bottomrule
    \toprule
    $K$ & 10 & 20 & 30 & 50 & 100\\
    \midrule
    Avg. DC $\uparrow$ & 87.95 & \textbf{88.17} & 88.11 & 87.80 & 87.13\\
    \bottomrule
\end{tabular}}
\label{delta_k}
\end{table}

We conduct experiments on Fundus dataset with only 20 labeled data. In Tab.~\ref{fundus}, the results for optic cup and disc are separated by a slash. The results confirm the significance of knowledge transfer and diverse intermediate samples. For other methods, severe error accumulation leads to poor performance. UST-RUN achieves outstanding results, with an average DC at least 4.70\% higher than other methods. When the number of labeled data is reduced to 10, UST-RUN experiences only a slight performance drop and still maintains a significant advantage compared to other methods with 40 labeled data.
As shown in Fig.~\ref{fundus_img}, UST-RUN achieves the best alignment with the ground truth in terms of segmentation results.
We integrate various domain adaptation methods with the SSMIS approach for a fair comparison. Utilizing FixMatch as a foundation, we select FDA~\cite{yang2020fda}, CutMix~\cite{yun2019cutmix}, ClassMix~\cite{olsson2021classmix}, CowMix~\cite{french2020milking}, and FMix~\cite{harris2020fmix} to facilitate knowledge transfer. Specifically, the FDA involves style transfer from labeled to unlabeled data, while other methods blend images using masks of different shapes. The results on the Fundus dataset are presented in Tab.~\ref{fundus_ssmsuda}. By constructing a more diverse set of intermediate samples and designing a more efficient training strategy for hard samples, UST-RUN achieves further performance improvements.

\subsubsection{Results on Prostate dataset} 
As shown in Tab.~\ref{prostate}, on Prostate dataset, UST-RUN demonstrates notable advantage and increased robustness when trained with 40 labeled data. Experiments with 20 labeled data show that UST-RUN exhibits minor performance degradation compared to UST.
Visual comparisons on Prostate dataset are presented in Fig.~\ref{prostate_img}.

\subsubsection{Results on M\&Ms dataset} 
In Tab.~\ref{mnms}, UST-RUN consistently outperforms other methods using 20 labeled data on M\&Ms dataset. The notable gap between UST-RUN and the upper bound can be attributed to a large number of labeled data. When training with all data from a single domain as labeled data, the number of labeled data far exceeds the 20 samples we utilized in other experiments. \blue{Visual comparisons on M\&Ms dataset are presented in Fig.~\ref{mnms_img}.}

\subsubsection{Results on BUSI dataset} 
As shown in Tab.~\ref{busi}, UST achieves the best performance with 64 and 129 labeled data. Building on this, UST-RUN further amplifies the advantages, improving the Dice coefficient (DC) by 1.93\% and 2.19\%. \blue{Visual comparisons on BUSI dataset are presented in Fig.~\ref{busi_img}}.

\subsubsection{Discussions}
\blue{In the MiDSS setting, the key challenge is to ensure the model's performance in the domain of labeled data while effectively transferring domain knowledge to unlabeled data through intermediate samples. Our method simultaneously considers the diversity of intermediate samples and the quality of pseudo-labels generated for both unlabeled and intermediate samples.}
\blue{Regarding existing approaches, SSMIS methods struggle to address domain shifts among samples, where low-quality pseudo-labels exacerbate error accumulation during self-training. Proper data perturbation can help simulate and explore distributions from other domains, while effective pseudo-label denoising strategies facilitate stable self-training. This explains why methods such as FixMatch and BCP demonstrate relatively commendable performance.
UDA methods typically rely on a large number of labeled source domain samples to provide sufficient domain information. However, this requirement is not met in the SSMIS setting, limiting the effectiveness of domain knowledge transfer.
Moreover, in the MiDSS setting, methods based on fine-tuning a foundational model are prone to overfitting on the limited labeled data from a single domain, potentially compromising their generalization to other domains.}

\blue{Our UST-RUN achieves the best performance across all four datasets.} More importantly, compared to UST, UST-RUN achieves further advancements and demonstrates greater stability with fewer annotations.

\subsection{Ablation Study}
\label{subsec::ablation_study}

\begin{figure}[!t]
\centering
\includegraphics[width=\columnwidth]{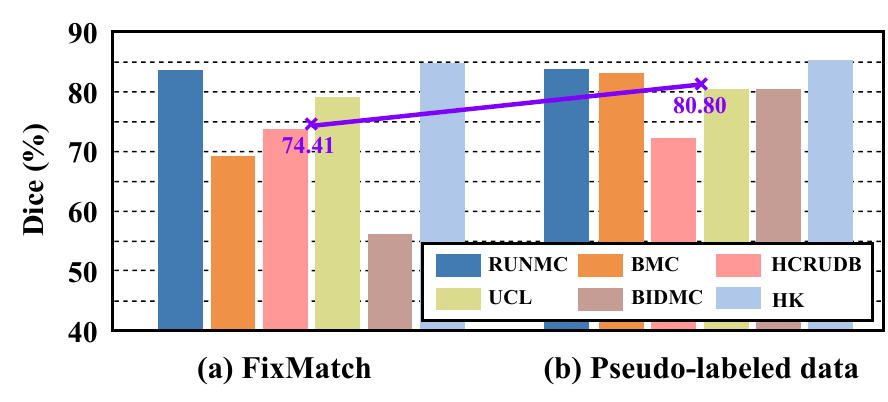}
\caption{Ablation experiments on the Prostate dataset directly utilizing pseudo-labeled data without CutMix. $\times$ represents the average DC across six domains for both methods.}
\label{fig:woCutMix}
\end{figure}

\begin{table}[h]
\centering
\footnotesize
\caption{\bluee{Influence of different pasting region selection strategies on Prostate dataset. Label-Guided, Pseudo-Guided, and Dual-Guided refer to pasting regions determined based on ground truth labels, pseudo labels, and both, respectively.}}
\resizebox{0.9\linewidth}{!}{
\begin{tabular}{lccccc}
    \toprule
    \multirow{2}{*}{Method} & \multirow{2}{*}{\#L} & DC $\uparrow$ & JC $\uparrow$ & HD $\downarrow$ & ASD $\downarrow$\\
    \cmidrule{3-6}
    & & \multicolumn{4}{c}{Avg.}\\
    \midrule
    Label-Guided & 20 & 81.49 & 72.73 & 38.27 & 21.46\\
    Pseudo-Guided & 20 & 83.30 & 75.51 & 15.57 & 8.46\\
    Dual-Guided & 20 & 84.16 & 75.66 & 16.42 & 7.01\\
    UST-RUN & 20 & \textbf{84.79} & \textbf{76.28} & \textbf{15.46} & \textbf{6.48}\\
    \bottomrule
\end{tabular}
}
\label{tab:pasting_region}
\end{table}
\begin{table}[t]
\centering
\footnotesize
\caption{\blue{Influence of randomly selected reliable and unreliable samples on Prostate dataset.}}
\resizebox{0.9\linewidth}{!}{
\begin{tabular}{lccccc}
    \toprule
    \multirow{2}{*}{\blue{Method}} & \multirow{2}{*}{\blue{\#L}} & \blue{DC $\uparrow$} & \blue{JC $\uparrow$} & \blue{HD $\downarrow$} & \blue{ASD $\downarrow$}\\
    \cmidrule{3-6}
    & & \multicolumn{4}{c}{\blue{Avg.}}\\
    \midrule
    \blue{Rand-RUN} & \blue{20} & \blue{77.07} & \blue{67.74} & \blue{49.50} & \blue{22.06}\\
    \blue{Rand-R} & \blue{20} & \blue{81.16} & \blue{72.40} & \blue{30.23} & \blue{13.74}\\
    \blue{Rand-UN} & \blue{20} & \blue{83.99} & \blue{75.64} & \blue{16.88} & \blue{6.99}\\
    \blue{UST-RUN} & \blue{20} & \blue{\textbf{84.79}} & \blue{\textbf{76.28}} & \blue{\textbf{15.46}} & \blue{\textbf{6.48}}\\
    \bottomrule
\end{tabular}}
\label{tab:random_run}
\end{table}

\begin{figure*}[!t]
\centering
\subfloat[][\footnotesize Dice analysis]{\includegraphics[width=0.22\linewidth]{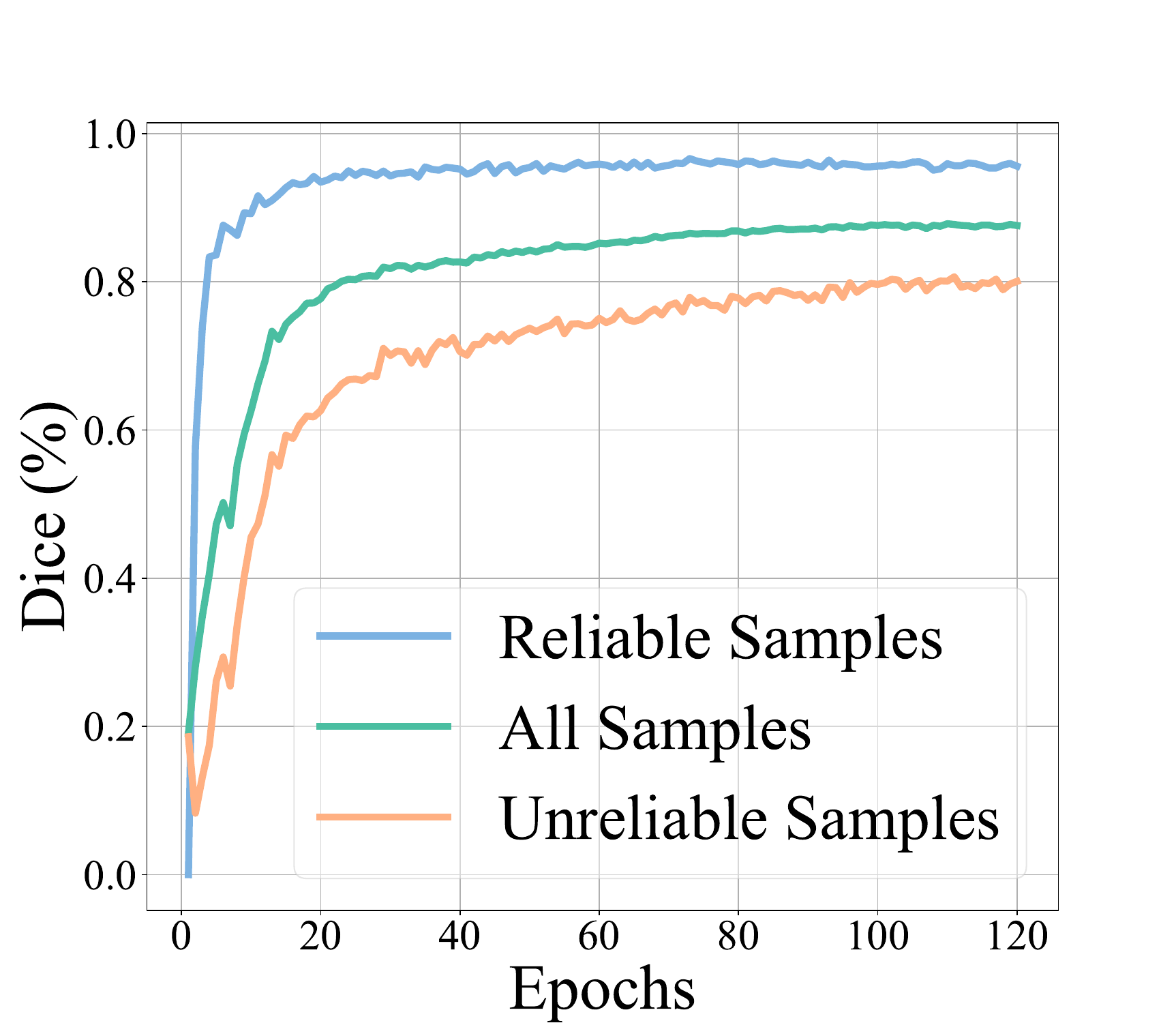}%
\label{re_vs_all_va_unre}}
\subfloat[][\footnotesize Number distribution]{\includegraphics[width=0.22\linewidth]{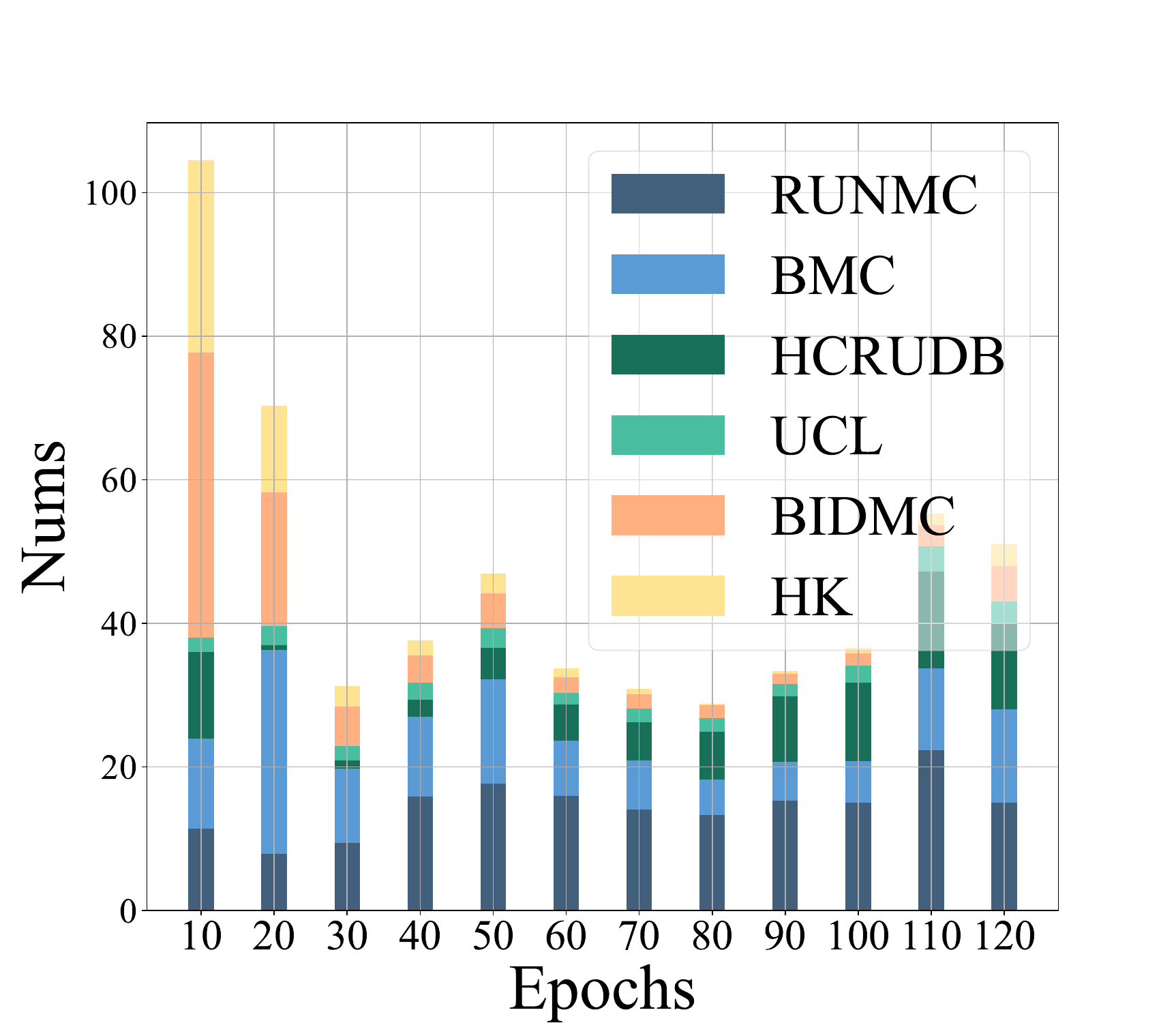}%
\label{reliable_ulb_num}}
\subfloat[][\footnotesize Selection frequency] {\includegraphics[width=0.22\linewidth]{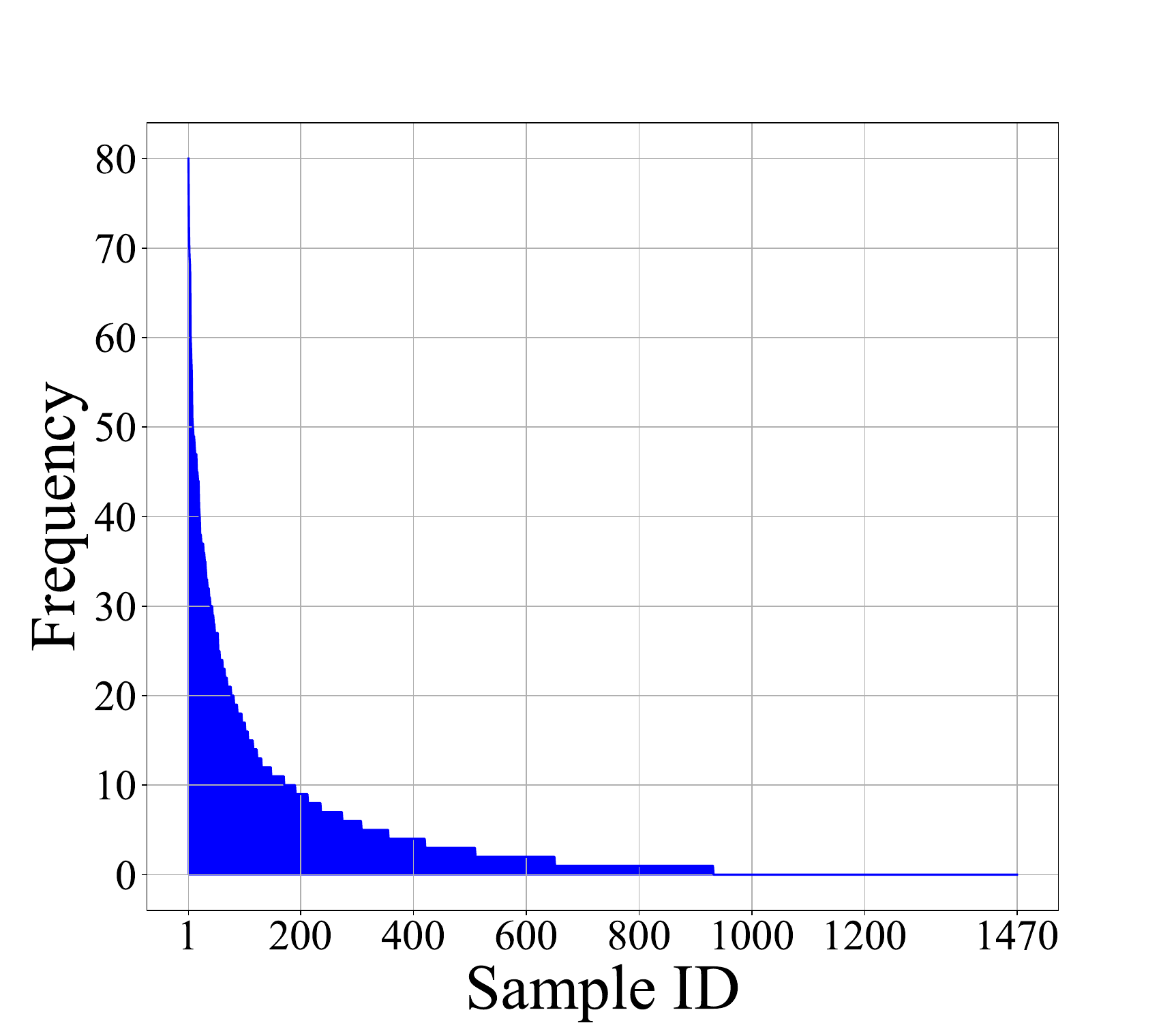}%
\label{frequency}}
\subfloat[][\footnotesize Hardness threshold]{\includegraphics[width=0.22\linewidth]{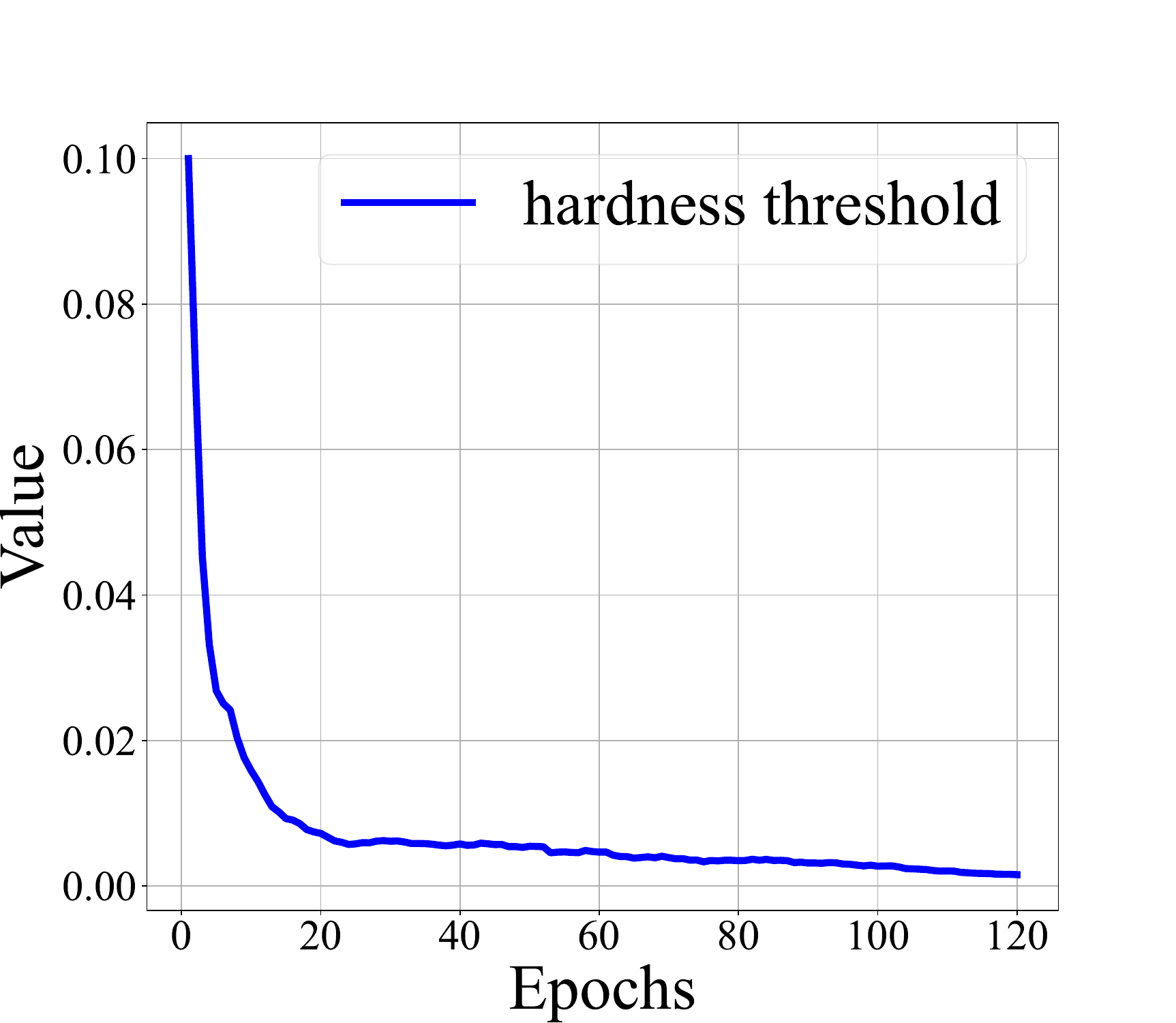}%
\label{hardness_threshold}}
\caption{Further analysis of reliable samples. The results are obtained from an experiment using 40 labeled data from the Prostate Domain BIDMC and the rest data as unlabeled data. (a) shows the average Dice coefficient of reliable samples, all unlabeled samples, and unreliable samples in each epoch. (b) reports the average number of reliable samples every 10 epochs. (c) demonstrates the frequency of each unlabeled data being selected as a reliable sample throughout the entire training process. (d) shows the variation tendency of the hardness threshold during model training.}
\label{fig_sim}
\end{figure*}

\subsubsection{The Efficiency of Each Module}
We conduct several experiments on Fundus dataset to validate the effectiveness of each component.
The results are shown in Tab.~\ref{fundusablation}. Compared to \#1, \#3 effectively leverages the intermediate domain information, aiding the model in making more accurate predictions for unlabeled data. In \#2, the straightforward guidance from $\hat{p}$ to $p^s$ has a limited impact.
In \#4, compared to \#5, the gradual construction of intermediate domains facilitates the transfer of domain knowledge.
In \#6, we incorporate two modules and achieve better performance. In \#7, we select reliable samples to generate diverse intermediate samples between any two domains, leading to a further improvement. In \#8, we further perform additional training on the unreliable samples, ultimately achieving an average performance of 88.15\%.

\subsubsection{Parameter Sensitivity}
\label{subsubsec::Parameter}
As shown in Fig.~\ref{beta_tau}, on Fundus dataset with 20 labeled data, we explore the relationship between size of low-frequency regions, confidence threshold, and model performance. Here, $M_\beta$ represents the low-frequency region of an image with a size of $2\beta W \times 2\beta H$, $\beta\in\{0.2,0.1,0.005,0.01,0.005\}$. There is a slight trend of performance improvement followed by a decline, with the peak achieved when $\beta=0.01$. Excessively large fusion areas may introduce high-frequency information, altering the geometric details of labeled data, while too small areas limit the integration of low-frequency information, leading to insufficient stylistic blending. For threshold $\tau$, the value of 0.95 works best, and other values do not significantly affect performance.
We also conduct ablation experiments of increasing coefficients $\beta$ and Queue capacity $K$ on Prostate dataset with 40 labeled data. Specifically, we set $\beta$ as $1+\rho$, $\rho \in \{1\times 10^{-4},5\times 10^{-4},1\times 10^{-3},5\times 10^{-3},1\times 10^{-2}\}$. Similarly, we set $K$ as 10, 20, 30, 50 and 100. In Tab.~\ref{delta_k}, we observe that the model performs optimally when $\delta$ is 1.0005 and $K$ is 20. The experimental results align with our intuition. Larger values of $\delta$ and $K$ imply that $\gamma$ is more likely to take larger values, meaning that the selection criteria for reliable samples are relaxed, resulting in lower quality pseudo-labels for reliable samples. However, small values make the criteria overly strict, making it challenging to select adequate reliable samples for generating diverse intermediate samples.

\subsubsection{Comparison with other pasting strategies}
\bluee{Randomly selected pasting regions may disrupt the semantic continuity of segmentation targets. To further investigate the role of semantic continuity, we designed alternative strategies to preserve the semantic structure of the pasted regions. These include selecting regions that (1) fully contain the ground-truth object from labeled data (\textit{Label-Guided}), (2) fully cover pseudo-labeled objects from unlabeled data (\textit{Pseudo-Guided}), or (3) include both (\textit{Dual-Guided}). These strategies are designed to preserve the semantic continuity of objects in labeled, unlabeled, or both types of samples, respectively. Experimental results (see Tab.~\ref{tab:pasting_region}) show that our default random region selection strategy yields the best performance. Introducing semantic discontinuity actually improves segmentation robustness by increasing the diversity of the generated intermediate samples. In contrast, enforcing semantic continuity significantly reduces the diversity of intermediate samples, leading to sub-optimal results. Moreover, our SymGD module is explicitly designed to break the semantic continuity of unlabeled data and integrate the separated information from paired intermediate samples. This mechanism provides supervision signals from the intermediate-domain perspective, thereby further improving the quality of pseudo labels.}

\subsubsection{Randomly selected reliable and unreliable samples}
\blue{To demonstrate that our criterion can effectively select ``real'' reliable samples, we conduct additional experiments by randomly select reliable and unreliable samples. As shown in Tab.~\ref{tab:random_run}, when the number of labeled data is 20 in Prostate dataset, randomly selecting either reliable (\textit{Rand-R}) or unreliable (\textit{Rand-UN}) samples negatively impacts the overall performance of our method. Moreover, when both are randomly relected(\textit{Rand-RUN}), the performance significantly lags behind that of the full RUN-UST.}

\subsubsection{The Utilization of Pseudo-labeled Data without CutMix} 
\blue{Due to the capability of reliable samples from multiple domains to provide domain-specific knowledge for unlabeled data, we directly treat these reliable samples as pseudo-labeled data (with confidence maps not being uniformly one-hot distributions) based on FixMatch framework. Specifically, we conduct the experiment on the Prostate dataset with 40 labeled data and do not apply CutMix for knowledge transfer. These reliable samples are incorporated into the calculation of the supervised loss during the training stage. The correspongding experimental results are displayed in Fig.~\ref{fig:woCutMix} under the term `Pseudo-labeled data'. Compared to FixMatch, reliable samples contribute to achieving superior model performance but still fall below our overall method. This not only validates the value of the reliable samples but also underscores the importance of CutMix.}

\subsubsection{Further Analysis of Reliable Samples}
\begin{figure}[!t]
\centering
\includegraphics[width=0.83\linewidth]{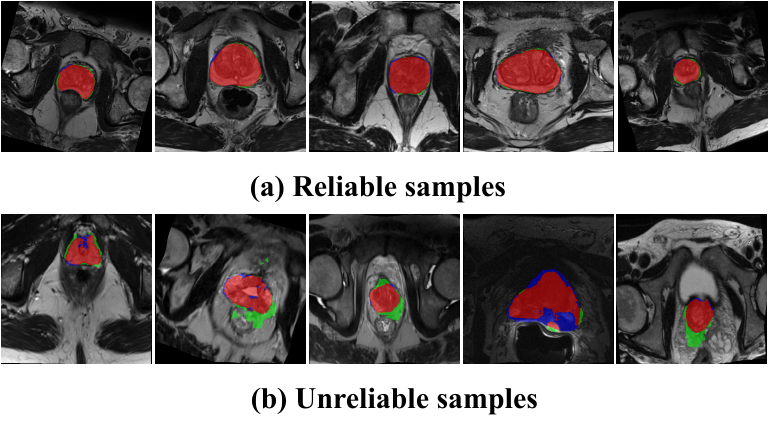}
\caption{Examples of reliable samples and unreliable samples. Red indicates true positive regions, green indicates false positive regions, and blue indicates false negative regions. }
\label{re_unre}
\end{figure}

We conduct further analysis of reliable samples on Prostate dataset. As shown in Fig.~\ref{re_vs_all_va_unre}, we compare the average DC of reliable samples, all unlabeled samples, and unreliable samples at each epoch, represented by blue, green, and orange lines, respectively. We observe that the average DC of reliable samples consistently surpassed that of all unlabeled samples, and the average DC of all unlabeled samples consistently exceeded that of unreliable samples. Fig.~\ref{reliable_ulb_num} displays the average number of reliable samples every 10 epochs, and around 40 reliable samples are selected per epoch. We count the frequency each unlabeled samples is selected as a reliable sample throughout the entire training. We assigned a unique identifier to each unlabeled samples based on its selection frequency, as shown in Fig.~\ref{frequency}. Among 1,470 unlabeled samples, the maximum frequency is 80, with approximately 200 samples being selected more than 10 times, and about 500 samples not selected at all. In Fig.~\ref{re_unre}, we illustrate some reliable and unreliable samples selected during training. Reliable samples show minimal differences from the ground truth, while unreliable samples exhibit the opposite characteristics. Moreover, we track the variation tendency of hardness threshold $\gamma$, which is shown in Fig.~\ref{hardness_threshold}. The fluctuations indicate that the criterion is dynamically adjusted, ensuring both the quantity and quality of reliable samples.

\section{Conclusion}
\label{conclusion}
In this paper, we investigate a novel and practical scenario, MiDSS, which involves the dual  challenges of limited annotation and domain shift.
To address the issue of declining pseudo-label quality due to domain shift, we construct intermediate domains by UCP. Besides, we introduce SymGD to enhance the utilization of intermediate domains information. Considering the stylistic differences between different domains, we design TP-RAM to introduce comprehensive and stable style transition components to intermediate domains. Building upon this, to further enhance diversity, we select and leverage reliable samples to generate intermediate samples. To achieve efficient knowledge transfer for unreliable samples, we additionally generate intermediate samples with high-quality pseudo-labels. Extensive experimental results on four datasets demonstrate the effectiveness of UST-RUN, showcasing significant improvements compared to existing methods.
\bibliographystyle{IEEEtran}
\bibliography{egbib}

\begin{thebibliography}{10}
\providecommand{\url}[1]{#1}
\csname url@samestyle\endcsname
\providecommand{\newblock}{\relax}
\providecommand{\bibinfo}[2]{#2}
\providecommand{\BIBentrySTDinterwordspacing}{\spaceskip=0pt\relax}
\providecommand{\BIBentryALTinterwordstretchfactor}{4}
\providecommand{\BIBentryALTinterwordspacing}{\spaceskip=\fontdimen2\font plus
\BIBentryALTinterwordstretchfactor\fontdimen3\font minus \fontdimen4\font\relax}
\providecommand{\BIBforeignlanguage}[2]{{%
\expandafter\ifx\csname l@#1\endcsname\relax
\typeout{** WARNING: IEEEtran.bst: No hyphenation pattern has been}%
\typeout{** loaded for the language `#1'. Using the pattern for}%
\typeout{** the default language instead.}%
\else
\language=\csname l@#1\endcsname
\fi
#2}}
\providecommand{\BIBdecl}{\relax}
\BIBdecl

\bibitem{liu2020shape}
Q.~Liu, Q.~Dou, and P.-A. Heng, ``Shape-aware meta-learning for generalizing prostate mri segmentation to unseen domains,'' in \emph{MICCAI}.\hskip 1em plus 0.5em minus 0.4em\relax Springer, 2020, pp. 475--485.

\bibitem{yu2019uncertainty}
L.~Yu, S.~Wang, X.~Li, C.-W. Fu, and P.-A. Heng, ``Uncertainty-aware self-ensembling model for semi-supervised 3d left atrium segmentation,'' in \emph{MICCAI}.\hskip 1em plus 0.5em minus 0.4em\relax Springer, 2019, pp. 605--613.

\bibitem{fan2020inf}
D.-P. Fan, T.~Zhou, G.-P. Ji, Y.~Zhou, G.~Chen, H.~Fu, J.~Shen, and L.~Shao, ``Inf-net: Automatic covid-19 lung infection segmentation from ct images,'' \emph{TMI}, 2020.

\bibitem{li2020transformation}
X.~Li, L.~Yu, H.~Chen, C.-W. Fu, L.~Xing, and P.-A. Heng, ``Transformation-consistent self-ensembling model for semisupervised medical image segmentation,'' \emph{TNNLS}, 2020.

\bibitem{zeng2023ss}
L.-L. Zeng, K.~Gao, D.~Hu, Z.~Feng, C.~Hou, P.~Rong, and W.~Wang, ``Ss-tbn: A semi-supervised tri-branch network for covid-19 screening and lesion segmentation,'' \emph{TPAMI}, 2023.

\bibitem{he2020momentum}
K.~He, H.~Fan, Y.~Wu, S.~Xie, and R.~Girshick, ``Momentum contrast for unsupervised visual representation learning,'' in \emph{CVPR}, 2020.

\bibitem{dash2019big}
S.~Dash, S.~K. Shakyawar, M.~Sharma, and S.~Kaushik, ``Big data in healthcare: management, analysis and future prospects,'' \emph{Journal of big data}, vol.~6, no.~1, pp. 1--25, 2019.

\bibitem{rudrapatna2020opportunities}
V.~A. Rudrapatna, A.~J. Butte \emph{et~al.}, ``Opportunities and challenges in using real-world data for health care,'' \emph{The Journal of Clinical Investigation}, 2020.

\bibitem{wang2019semi}
Q.~Wang, W.~Li, and L.~V. Gool, ``Semi-supervised learning by augmented distribution alignment,'' in \emph{ICCV}, 2019.

\bibitem{guan2021domain}
H.~Guan and M.~Liu, ``Domain adaptation for medical image analysis: a survey,'' \emph{T-BME}, 2021.

\bibitem{chen2020unsupervised}
C.~Chen, Q.~Dou, H.~Chen, J.~Qin, and P.~A. Heng, ``Unsupervised bidirectional cross-modality adaptation via deeply synergistic image and feature alignment for medical image segmentation,'' \emph{TMI}, 2020.

\bibitem{russo2018source}
P.~Russo, F.~M. Carlucci, T.~Tommasi, and B.~Caputo, ``From source to target and back: symmetric bi-directional adaptive gan,'' in \emph{CVPR}, 2018.

\bibitem{dou2018unsupervised}
Q.~Dou, C.~Ouyang, C.~Chen, H.~Chen, and P.-A. Heng, ``Unsupervised cross-modality domain adaptation of convnets for biomedical image segmentations with adversarial loss,'' \emph{arXiv}, 2018.

\bibitem{li2021ecacl}
K.~Li, C.~Liu, H.~Zhao, Y.~Zhang, and Y.~Fu, ``Ecacl: A holistic framework for semi-supervised domain adaptation,'' in \emph{Proceedings of the IEEE/CVF international conference on computer vision}, 2021, pp. 8578--8587.

\bibitem{kim2020attract}
T.~Kim and C.~Kim, ``Attract, perturb, and explore: Learning a feature alignment network for semi-supervised domain adaptation,'' in \emph{European conference on computer vision}.

\bibitem{yan2022multi}
Z.~Yan, Y.~Wu, G.~Li, Y.~Qin, X.~Han, and S.~Cui, ``Multi-level consistency learning for semi-supervised domain adaptation,'' \emph{arXiv preprint arXiv:2205.04066}, 2022.

\bibitem{yun2019cutmix}
S.~Yun, D.~Han, S.~J. Oh, S.~Chun, J.~Choe, and Y.~Yoo, ``Cutmix: Regularization strategy to train strong classifiers with localizable features,'' in \emph{ICCV}, 2019.

\bibitem{wang2022separated}
J.~Wang, X.~Li, Y.~Han, J.~Qin, L.~Wang, and Z.~Qichao, ``Separated contrastive learning for organ-at-risk and gross-tumor-volume segmentation with limited annotation,'' in \emph{AAAI}, 2022.

\bibitem{liu2021feddg}
Q.~Liu, C.~Chen, J.~Qin, Q.~Dou, and P.-A. Heng, ``Feddg: Federated domain generalization on medical image segmentation via episodic learning in continuous frequency space,'' in \emph{CVPR}, 2021.

\bibitem{ma2024constructing}
Q.~Ma, J.~Zhang, L.~Qi, Q.~Yu, Y.~Shi, and Y.~Gao, ``Constructing and exploring intermediate domains in mixed domain semi-supervised medical image segmentation,'' in \emph{CVPR}, 2024.

\bibitem{zhuang2013challenges}
X.~Zhuang, ``Challenges and methodologies of fully automatic whole heart segmentation: a review,'' \emph{Journal of Healthcare Engineering}, 2013.

\bibitem{li2020shape}
S.~Li, C.~Zhang, and X.~He, ``Shape-aware semi-supervised 3d semantic segmentation for medical images,'' in \emph{MICCAI}, 2020.

\bibitem{luo2021semi}
X.~Luo, J.~Chen, T.~Song, and G.~Wang, ``Semi-supervised medical image segmentation through dual-task consistency,'' in \emph{AAAI}, 2021.

\bibitem{wu2022exploring}
Y.~Wu, Z.~Wu, Q.~Wu, Z.~Ge, and J.~Cai, ``Exploring smoothness and class-separation for semi-supervised medical image segmentation,'' in \emph{MICCAI}, 2022.

\bibitem{miao2023caussl}
J.~Miao, C.~Chen, F.~Liu, H.~Wei, and P.-A. Heng, ``Caussl: Causality-inspired semi-supervised learning for medical image segmentation,'' in \emph{ICCV}, 2023.

\bibitem{chi2024adaptive}
H.~Chi, J.~Pang, B.~Zhang, and W.~Liu, ``Adaptive bidirectional displacement for semi-supervised medical image segmentation,'' in \emph{Proceedings of the IEEE/CVF conference on computer vision and pattern recognition}, 2024, pp. 4070--4080.

\bibitem{kirillov2023segment}
A.~Kirillov, E.~Mintun, N.~Ravi, H.~Mao, C.~Rolland, L.~Gustafson, T.~Xiao, S.~Whitehead, A.~C. Berg, W.-Y. Lo \emph{et~al.}, ``Segment anything,'' in \emph{ICCV}, 2023, pp. 4015--4026.

\bibitem{zhang2023customized}
K.~Zhang and D.~Liu, ``Customized segment anything model for medical image segmentation,'' \emph{arXiv}, 2023.

\bibitem{chen2023sam}
T.~Chen, L.~Zhu, C.~Ding, R.~Cao, Y.~Wang, Z.~Li, L.~Sun, P.~Mao, and Y.~Zang, ``Sam fails to segment anything?--sam-adapter: Adapting sam in underperformed scenes: Camouflage, shadow, medical image segmentation, and more,'' \emph{arXiv preprint arXiv:2304.09148}, 2023.

\bibitem{ma2024segment}
J.~Ma, Y.~He, F.~Li, L.~Han, C.~You, and B.~Wang, ``Segment anything in medical images,'' \emph{Nature Communications}, vol.~15, no.~1, p. 654, 2024.

\bibitem{cheng2024unleashing}
Z.~Cheng, Q.~Wei, H.~Zhu, Y.~Wang, L.~Qu, W.~Shao, and Y.~Zhou, ``Unleashing the potential of sam for medical adaptation via hierarchical decoding,'' in \emph{Proceedings of the IEEE/CVF Conference on Computer Vision and Pattern Recognition}, 2024, pp. 3511--3522.

\bibitem{chartsias2017adversarial}
A.~Chartsias, T.~Joyce, R.~Dharmakumar, and S.~A. Tsaftaris, ``Adversarial image synthesis for unpaired multi-modal cardiac data,'' in \emph{SASHIMI}, 2017.

\bibitem{ganin2016domain}
Y.~Ganin, E.~Ustinova, H.~Ajakan, P.~Germain, H.~Larochelle, F.~Laviolette, M.~Marchand, and V.~Lempitsky, ``Domain-adversarial training of neural networks,'' \emph{JMLR}, 2016.

\bibitem{zhao2022uda}
Z.~Zhao, F.~Zhou, K.~Xu, Z.~Zeng, C.~Guan, and S.~K. Zhou, ``Le-uda: Label-efficient unsupervised domain adaptation for medical image segmentation,'' \emph{TMI}, 2022.

\bibitem{tsai2018learning}
Y.-H. Tsai, W.-C. Hung, S.~Schulter, K.~Sohn, M.-H. Yang, and M.~Chandraker, ``Learning to adapt structured output space for semantic segmentation,'' in \emph{CVPR}, 2018.

\bibitem{zheng2021rectifying}
Z.~Zheng and Y.~Yang, ``Rectifying pseudo label learning via uncertainty estimation for domain adaptive semantic segmentation,'' \emph{IJCV}, 2021.

\bibitem{zhang2020collaborative}
Y.~Zhang, Y.~Wei, Q.~Wu, P.~Zhao, S.~Niu, J.~Huang, and M.~Tan, ``Collaborative unsupervised domain adaptation for medical image diagnosis,'' \emph{TIP}, 2020.

\bibitem{chen2022deliberated}
L.~Chen, Z.~Wei, X.~Jin, H.~Chen, M.~Zheng, K.~Chen, and Y.~Jin, ``Deliberated domain bridging for domain adaptive semantic segmentation,'' \emph{NeurIPS}, vol.~35, pp. 15\,105--15\,118, 2022.

\bibitem{yang2020fda}
Y.~Yang and S.~Soatto, ``Fda: Fourier domain adaptation for semantic segmentation,'' in \emph{CVPR}, 2020.

\bibitem{wang2020tent}
D.~Wang, E.~Shelhamer, S.~Liu, B.~Olshausen, and T.~Darrell, ``Tent: Fully test-time adaptation by entropy minimization,'' \emph{arXiv}, 2020.

\bibitem{chen2024each}
Z.~Chen, Y.~Pan, Y.~Ye, M.~Lu, and Y.~Xia, ``Each test image deserves a specific prompt: Continual test-time adaptation for 2d medical image segmentation,'' in \emph{CVPR}, 2024.

\bibitem{yi2023source}
L.~Yi, G.~Xu, P.~Xu, J.~Li, R.~Pu, C.~Ling, A.~I. McLeod, and B.~Wang, ``When source-free domain adaptation meets learning with noisy labels,'' \emph{arXiv preprint arXiv:2301.13381}, 2023.

\bibitem{singh2021improving}
A.~Singh, N.~Doraiswamy, S.~Takamuku, M.~Bhalerao, T.~Dutta, S.~Biswas, A.~Chepuri, B.~Vengatesan, and N.~Natori, ``Improving semi-supervised domain adaptation using effective target selection and semantics,'' in \emph{Proceedings of the IEEE/CVF conference on computer vision and pattern recognition}, 2021, pp. 2709--2718.

\bibitem{qin2021contradictory}
C.~Qin, L.~Wang, Q.~Ma, Y.~Yin, H.~Wang, and Y.~Fu, ``Contradictory structure learning for semi-supervised domain adaptation,'' in \emph{Proceedings of the 2021 SIAM International Conference on Data Mining (SDM)}.\hskip 1em plus 0.5em minus 0.4em\relax SIAM, 2021, pp. 576--584.

\bibitem{jiang2020bidirectional}
P.~Jiang, A.~Wu, Y.~Han, Y.~Shao, M.~Qi, and B.~Li, ``Bidirectional adversarial training for semi-supervised domain adaptation.'' in \emph{IJCAI}, 2020, pp. 934--940.

\bibitem{huang2023semi}
X.~Huang, C.~Zhu, and W.~Chen, ``Semi-supervised domain adaptation via prototype-based multi-level learning,'' \emph{arXiv preprint arXiv:2305.02693}, 2023.

\bibitem{zhang2017mixup}
H.~Zhang, M.~Cisse, Y.~N. Dauphin, and D.~Lopez-Paz, ``mixup: Beyond empirical risk minimization,'' \emph{arXiv}, 2017.

\bibitem{bai2023bidirectional}
Y.~Bai, D.~Chen, Q.~Li, W.~Shen, and Y.~Wang, ``Bidirectional copy-paste for semi-supervised medical image segmentation,'' in \emph{CVPR}, 2023.

\bibitem{fan2022ucc}
J.~Fan, B.~Gao, H.~Jin, and L.~Jiang, ``Ucc: Uncertainty guided cross-head co-training for semi-supervised semantic segmentation,'' in \emph{CVPR}, 2022.

\bibitem{tarvainen2017mean}
A.~Tarvainen and H.~Valpola, ``Mean teachers are better role models: Weight-averaged consistency targets improve semi-supervised deep learning results,'' \emph{NeurIPS}, vol.~30, pp. 1195--1204, 2017.

\bibitem{sohn2020fixmatch}
K.~Sohn, D.~Berthelot, N.~Carlini, Z.~Zhang, H.~Zhang, C.~A. Raffel, E.~D. Cubuk, A.~Kurakin, and C.-L. Li, ``Fixmatch: Simplifying semi-supervised learning with consistency and confidence,'' \emph{NeurIPS}, vol.~33, pp. 596--608, 2020.

\bibitem{upretee2022fixmatchseg}
P.~Upretee and B.~Khanal, ``Fixmatchseg: Fixing fixmatch for semi-supervised semantic segmentation,'' \emph{arXiv}, 2022.

\bibitem{yang2023revisiting}
L.~Yang, L.~Qi, L.~Feng, W.~Zhang, and Y.~Shi, ``Revisiting weak-to-strong consistency in semi-supervised semantic segmentation,'' in \emph{CVPR}, 2023.

\bibitem{grill2020bootstrap}
J.-B. Grill, F.~Strub, F.~Altch{\'e}, C.~Tallec, P.~Richemond, E.~Buchatskaya, C.~Doersch, B.~Avila~Pires, Z.~Guo, M.~Gheshlaghi~Azar \emph{et~al.}, ``Bootstrap your own latent-a new approach to self-supervised learning,'' \emph{NeurIPS}, vol.~33, pp. 21\,271--21\,284, 2020.

\bibitem{Wang_Xu_Liu_Zhang_Fu_2020}
Y.~Wang, C.~Xu, C.~Liu, L.~Zhang, and Y.~Fu, ``Instance credibility inference for few-shot learning,'' in \emph{CVPR}, 2020.

\bibitem{ronneberger2015u}
O.~Ronneberger, P.~Fischer, and T.~Brox, ``U-net: Convolutional networks for biomedical image segmentation,'' in \emph{MICCAI}, 2015.

\bibitem{wang2020dofe}
S.~Wang, L.~Yu, K.~Li, X.~Yang, C.-W. Fu, and P.-A. Heng, ``Dofe: Domain-oriented feature embedding for generalizable fundus image segmentation on unseen datasets,'' \emph{TMI}, 2020.

\bibitem{campello2021multi}
V.~M. Campello, P.~Gkontra, C.~Izquierdo, C.~Martin-Isla, A.~Sojoudi, P.~M. Full, K.~Maier-Hein, Y.~Zhang, Z.~He, J.~Ma \emph{et~al.}, ``Multi-centre, multi-vendor and multi-disease cardiac segmentation: the m\&ms challenge,'' \emph{TMI}, 2021.

\bibitem{al2020dataset}
W.~Al-Dhabyani, M.~Gomaa, H.~Khaled, and A.~Fahmy, ``Dataset of breast ultrasound images,'' \emph{Data in brief}, 2020.

\bibitem{lu2022unsupervised}
C.~Lu, S.~Zheng, and G.~Gupta, ``Unsupervised domain adaptation for cardiac segmentation: Towards structure mutual information maximization,'' in \emph{CVPR}, 2022.

\bibitem{olsson2021classmix}
V.~Olsson, W.~Tranheden, J.~Pinto, and L.~Svensson, ``Classmix: Segmentation-based data augmentation for semi-supervised learning,'' in \emph{WACV}, 2021.

\bibitem{french2020milking}
G.~French, A.~Oliver, and T.~Salimans, ``Milking cowmask for semi-supervised image classification,'' \emph{arXiv}, 2020.

\bibitem{harris2020fmix}
E.~Harris, A.~Marcu, M.~Painter, M.~Niranjan, A.~Pr{\"u}gel-Bennett, and J.~Hare, ``Fmix: Enhancing mixed sample data augmentation,'' \emph{arXiv}, 2020.

\end{thebibliography}

\end{document}